\documentclass[table,10pt]{article} 
\usepackage[accepted]{tmlr}

\usepackage{amsmath,amsfonts,bm}

\def\eqref#1{equation~\ref{#1}}

\def\1{\bm{1}}

\DeclareMathAlphabet{\mathsfit}{\encodingdefault}{\sfdefault}{m}{sl}
\SetMathAlphabet{\mathsfit}{bold}{\encodingdefault}{\sfdefault}{bx}{n}

\usepackage[utf8]{inputenc} 
\usepackage[T1]{fontenc}    
\usepackage{hyperref}       
\usepackage{url}            
\usepackage{booktabs}       
\usepackage{amsfonts}       
\usepackage{nicefrac}       
\usepackage{microtype}      
\usepackage{xcolor}         

\RequirePackage{fancyhdr}
\RequirePackage{algorithmic}
\RequirePackage{natbib}
\RequirePackage{eso-pic} 
\RequirePackage{forloop}

\definecolor{mydarkblue}{rgb}{0,0.08,0.45}
\definecolor{myblue}{HTML}{3b75c3}
\definecolor{myred}{HTML}{E33222}
\definecolor{mygreen}{HTML}{438773}
\definecolor{mymaroon}{RGB}{142,27,19}
\definecolor{maroon}{HTML}{800000}
\definecolor{mycite}{cmyk}{0.55,1,0,0.15}
\definecolor{codeblue}{rgb}{0.25,0.5,0.5}
\definecolor{codekw}{rgb}{0.85, 0.18, 0.50}
\definecolor{codegreen}{rgb}{0,0.6,0}
\definecolor{codegray}{rgb}{0.5,0.5,0.5}
\definecolor{codepurple}{rgb}{0.58,0,0.82}
\definecolor{backcolour}{rgb}{0.95,0.95,0.92}
\hypersetup{
    colorlinks=true,
    citecolor=codeblue,
    linkcolor=codeblue,
    urlcolor=maroon
          }

\usepackage{microtype}
\usepackage{graphicx}
\usepackage{subfigure}
\usepackage{booktabs} 

\usepackage[ruled,vlined,linesnumbered]{algorithm2e}

\usepackage{amsmath}
\usepackage{amssymb}
\usepackage{mathtools}
\usepackage{amsthm}
\usepackage{multirow}
\usepackage{wrapfig}
\usepackage{enumitem}

\usepackage[capitalize,noabbrev,nameinlink]{cleveref}

\theoremstyle{plain}

\theoremstyle{definition}

\theoremstyle{remark}

\newcommand{\cora}{\texttt{Cora}\xspace}
\newcommand{\citeseer}{\texttt{Citeseer}\xspace}
\newcommand{\cs}{\texttt{CS}\xspace}
\newcommand{\physics}{\texttt{Physics}\xspace}
\newcommand{\computers}{\texttt{Computers}\xspace}
\newcommand{\photos}{\texttt{Photos}\xspace}
\newcommand{\igbtiny}{\texttt{IGB-100K}\xspace}
\newcommand{\igbsmall}{\texttt{IGB\-1M}\xspace}

\newcommand{\isolated}{Isolated\xspace}
\newcommand{\cold}{Low-degree\xspace}
\newcommand{\warm}{Warm\xspace}
\newcommand{\overall}{Overall\xspace}

\newcommand{\innerproduct}{inner product\xspace}

\newcommand{\ours}{\textsc{NodeDup}\xspace}
\newcommand{\ourslight}{\textsc{NodeDup(L)}\xspace}

\newcommand{\ms}[2]{{#1\tiny{$\pm$#2}}}

\usepackage[textsize=tiny]{todonotes}

\usepackage{etoc}

\makeatletter
\def\part{\par
   \addvspace{4ex}%
   \@afterindentfalse
   \secdef\@part\@spart}%
\def\@part[#1]#2{%
 \@ifnum{\c@secnumdepth >\m@ne}{%
        \refstepcounter{part}%
        \addcontentsline{toc}{part}{\thepart\hspace{1em}#1}%
 }{%
      \addcontentsline{toc}{part}{#1}%
 }%
   \nobreak
   \vskip 3ex
   \@afterheading
}%
\makeatother

\title{Node Duplication Improves Cold-start Link Prediction}

\author{\name Zhichun Guo$^{1}$\thanks{Corresponds to zcguo@uw.edu.}~~Tong Zhao$^{2}$~~Yozen Liu$^{2}$~~Kaiwen Dong$^{3}$~~William Shiao$^{4}$~~Mingxuan Ju$^{2}$\\Neil Shah$^{2}$~~Nitesh V. Chawla$^{3}$\\
\addr $^{1}$University of Washington~~$^{2}$Snap Inc.~~$^{3}$University of Notre Dame~~$^{4}$University of California, Riverside}

\def\month{08}  
\def\year{2025} 
\def\openreview{\url{https://openreview.net/forum?id=hIOTzz87N9}} 

\begin{document}

\maketitle

\vspace{-0.1in}
\begin{abstract}
\vspace{-0.1in}
Graph Neural Networks (GNNs) are prominent in graph machine learning and have shown state-of-the-art performance in Link Prediction (LP) tasks. Nonetheless, recent studies show that GNNs struggle to produce good results on low-degree nodes despite their overall strong performance. In practical applications of LP, like recommendation systems, improving performance on low-degree nodes is critical, as it amounts to tackling the cold-start problem of improving the experiences of users with few observed interactions. In this paper, we investigate improving GNNs' LP performance on low-degree nodes while preserving their performance on high-degree nodes and propose a simple yet surprisingly effective augmentation technique called \ours. Specifically, \ours duplicates low-degree nodes and creates links between nodes and their own duplicates before following the standard supervised LP training scheme. By leveraging a ``multi-view'' perspective for low-degree nodes, \ours shows significant LP performance improvements on low-degree nodes without compromising any performance on high-degree nodes. Additionally, as a plug-and-play augmentation module, \ours can be easily applied on existing GNNs with very light computational cost. Extensive experiments show that \ours achieves \textbf{38.49\%}, \textbf{13.34\%}, and \textbf{6.76\%} relative improvements on isolated, low-degree, and warm nodes, respectively, on average across all datasets compared to GNNs and the existing cold-start methods. 
\end{abstract}

\vspace{-0.2in}
\section{Introduction}
Link prediction (LP) is a fundamental task of graph-structured data~\citep{liben2007link, trouillon2016complex}, which aims to predict the likelihood of the links existing between two nodes in the network. It has wide-ranging real-world applications across different domains, such as friend recommendations in social media~\citep{sankar2021graph, tang2022friend, fan2022graph}, product recommendations in e-commerce platforms~\citep{ying2018graph, he2020lightgcn}, knowledge graph completion~\citep{li2023message, vashishth2020composition, zhang2020few}, and chemical interaction prediction~\citep{stanfield2017drug, kovacs2019network, yang2021safedrug}. 

\begin{wrapfigure}{r}{0.46\linewidth}
    \centering
    \vspace{-0.3in}
    \includegraphics[width=\linewidth]{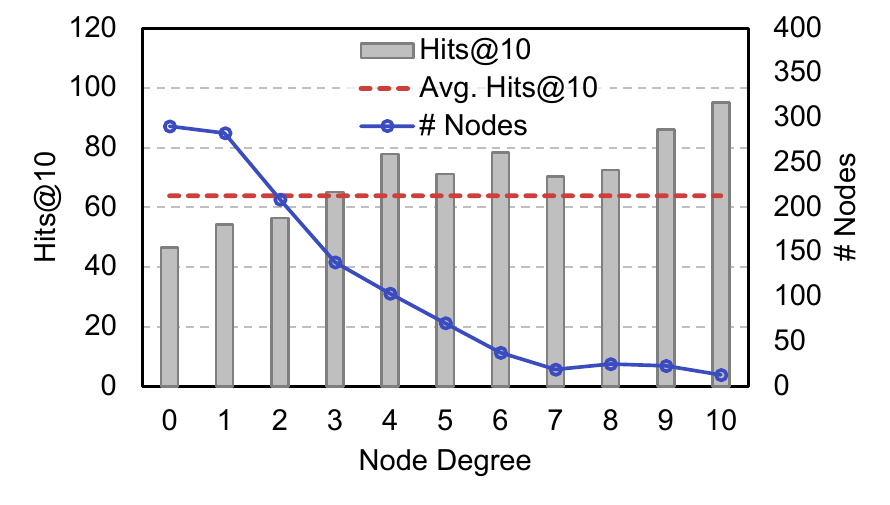}
    \label{fig:citeseer}
    \vspace{-0.3in}
    \caption{Node Degree Distribution and LP Performance (GSage as an encoder and \innerproduct as a decoder) Distribution w.r.t Nodes Degrees showing reverse trends on \citeseer dataset.}
    \label{fig:pre-analysis}
    \vspace{-0.3in}
\end{wrapfigure}

In recent years, graph neural networks (GNNs)~\citep{kipf2016semi, velivckovic2017graph, hamilton2017inductive} have been widely applied to LP, and a series of cutting-edge models have been proposed~\citep{zhang2018link, zhang2021labeling, zhu2021neural, zhao2022learning}. Most GNNs follow a message-passing scheme~\citep{gilmer2017neural} in which information is iteratively aggregated from neighbors and used to update node representations accordingly. Consequently, the success of GNNs usually heavily relies on having sufficient high-quality neighbors for each node~\citep{zheng2021cold, liu2021tail}. However, real-world graphs often exhibit long-tailed distribution in terms of node degrees, where a significant fraction of nodes have very few neighbors~\citep{tang2020investigating, ding2021zero, hao2021pre}. For example, \cref{fig:pre-analysis} shows the long-tailed degree distribution of the \citeseer dataset. Moreover, LP performances w.r.t. node degrees on this dataset also clearly indicate that GNNs struggle to generate satisfactory results for nodes with low or zero degrees. For simplicity, in this paper, we refer to the nodes with low or zero degrees as \emph{cold} nodes and the nodes with higher degrees as \emph{warm} nodes. 

To boost GNNs' performance on cold nodes, recent studies have proposed various training strategies~\citep{liu2020towards, liu2021tail, zheng2021cold, hu2022tuneup} and augmentation strategies~\citep{hu2022tuneup,rong2019dropedge,zhao2022learning} to improve representation learning quality. For instance, ColdBrew~\citep{zheng2021cold} posits that training a powerful MLP can rediscover missing neighbor information for cold nodes; TailGNN~\citep{liu2021tail} utilizes a cold-node-specific module to accomplish the same objective. However, such advanced training strategies (e.g., ColdBrew and TailGNN) share a notable drawback: they are trained with a bias towards cold nodes, which then sacrifices performance on warm nodes (empirically validated in \cref{tab:tailnodes}). However, in real-world applications, both cold nodes and warm nodes are critical~\citep{clauset2009power}. On the other hand, while augmentation methods such as LAGNN~\citep{liu2022local} do not have such bias, they primarily focus on improving the overall performance of GNNs in LP tasks, which may be dominated by warm nodes due to their higher connectivity. Additionally, the augmentation methods usually introduce a significant amount of extra computational costs (empirically validated in \cref{fig:aug}). In light of the existing work discussed above on improving LP performance for cold nodes, we are naturally motivated to explore the following crucial but rather unexplored research question:
\begin{center}
    \textbf{\textit{Can we improve LP performance on cold nodes without compromising warm node performance?}}
\end{center}

We observe that cold node LP performance usually suffers because they are under-represented in standard supervised LP training due to their few (if any) connections. Given this observation, in this work, we introduce a simple yet effective augmentation method, \ours, for improving LP performance on cold nodes. Specifically, \ours duplicates cold nodes and establishes edges between each original cold node and its corresponding duplicate. Subsequently, we conduct standard supervised end-to-end training of GNNs on the augmented graph. 
To better understand why \ours is able to improve LP performance for cold nodes, we thoroughly analyze it from multiple perspectives, during which we discover that this simple technique effectively offers a ``multi-view'' perspective of cold nodes during training. This ``multi-view'' perspective of the cold nodes acts similarly to an ensemble and drives performance improvements for these nodes. Additionally, our straightforward augmentation method provides valuable supervised training signals for cold nodes and especially isolated nodes. Furthermore, we also introduce \ourslight, a lightweight variation of \ours that adds only self-loop edges into training edges for cold nodes.
\ourslight empirically offers up to a 1.3$\times$ speedup over \ours for the training process and achieves significant speedup over existing augmentation baselines. In our experiments, we comprehensively evaluate our method on seven benchmark datasets. Compared to GNNs and the existing cold-start methods, \ours achieves \textbf{38.49\%}, \textbf{13.34\%}, and \textbf{6.76\%} relative improvements on isolated, low-degree, and warm nodes, respectively, on average across all datasets. \ours also greatly outperforms augmentation baselines on cold nodes, with comparable warm node performance. Finally, as plug-and-play augmentation methods, our methods are versatile and effective with different LP encoders/decoders. They also achieve significant performance in a more realistic inductive setting. Our code can be found at \url{https://github.com/zhichunguo/NodeDup}. 

\label{sec:introduction}

\vspace{-0.1in}
\section{Preliminaries}
\label{sec:preliminary}
\vspace{-0.1in}
\textbf{Notation.} Let an attributed graph be $G = \{\mathcal{V}, \mathcal{E}, \textbf{X}\}$, where $\mathcal{V}$ is the set of $N$ nodes and $\mathcal{E} \subseteq \mathcal{V} \times \mathcal{V}$ is the edges where each $e_{vu} \in \mathcal{E}$ indicates nodes $v$ and $u$ are linked. Let $\textbf{X} \in \mathbb{R} ^ {N \times F}$ be the node attribute matrix, where $F$ is the attribute dimension. Let $\mathcal{N}_v$ be the set of neighbors of node $v$, i.e., $\mathcal{N}_v = \{u | e_{vu} \in \mathcal{E}\}$, and the degree of node $v$ is $|\mathcal{N}_v|$. We separate the set of nodes $\mathcal{V}$ into three disjoint sets $\mathcal{V}_{iso}$, $\mathcal{V}_{low}$, and $\mathcal{V}_{warm}$ by their degrees based on the threshold hyperparameter $\delta$\footnote{This threshold $\delta$ is set as 2 in our experiments, based on observed performance gaps in LP on various datasets, as shown in \cref{fig:pre-analysis} and \cref{fig:pre-analysis-appendix}. Further reasons for this threshold are detailed in~\cref{sec:preanalysis_appendix}.}. For each node $v \in \mathcal{V}$, $v \in \mathcal{V}_{iso}$ if $|\mathcal{N}_v|=0$; $v \in \mathcal{V}_{low}$ if $0<|\mathcal{N}_v|\leq\delta$; $v \in \mathcal{V}_{warm}$ if $|\mathcal{N}_v|>\delta$. For ease of notation, we also use $\mathcal{V}_{cold}=\mathcal{V}_{iso}\cup\mathcal{V}_{low}$ to denote the cold nodes, which is the union of \isolated and \cold nodes.

\textbf{LP with GNNs.} In this work, we follow the commonly-used encoder-decoder framework for GNN-based LP~\citep{kipf2016variational,berg2017graph,schlichtkrull2018modeling,ying2018graph, davidson2018hyperspherical, zhu2021neural,yun2021neo,zhao2022learning}, where a GNN encoder learns the node representations and the decoder predicts the link existence probabilities given each pair of node representations. Most GNNs follow the message passing design~\citep{gilmer2017neural} that iteratively aggregate each node's neighbors' information to update its embeddings. Without the loss of generality, for each node $v$, the $l$-th layer of a GNN can be defined as 
\begin{equation}
\label{eq:gnn}
    \boldsymbol{h}_v^{(l)} = \text{\small UPDATE}\big(\boldsymbol{h}_v^{(l-1)}, \boldsymbol{m}_{v}^{(l-1)}\big),
    \text{s.t.} \;\; \boldsymbol{m}_{v}^{(l-1)} = \text{AGG}\big(\{\boldsymbol{h}_u^{(l-1)}\}: \forall u \in \mathcal{N}_v\big), 
\end{equation}
where $\boldsymbol{h}_v^{(l)}$ is the $l$-th layer's output representation of node $v$, $\boldsymbol{h}_v^{(0)} = \boldsymbol{x_v}$, $\text{\small AGG}(\cdot)$ is the (typically permutation-invariant) aggregation function, and $\text{\small UPDATE}(\cdot)$ is the update function that combines node $v$'s neighbor embedding and its own embedding from the previous layer. For any node pair $v$ and $u$, the decoding process can be defined as $\hat{y}_{vu} = \sigma\big(\text{\small DECODER}(\boldsymbol{h}_v, \boldsymbol{h}_u)\big),$
where $\boldsymbol{h}_v$ is the GNN's output representation for node $v$ and $\sigma$ is the Sigmoid function. Following existing literature, we use \innerproduct~\citep{wang2021pairwise,zheng2021cold} as the default $\text{\small DECODER}$.

The standard supervised LP training optimizes model parameters w.r.t. a training set, which is usually the union of all observed $M$ edges and $KM$ no-edge node pairs (as training with all $O(N^2)$ no-edges is infeasible in practice), where $K$ is the negative sampling rate ($K=1$ usually). We use $\mathcal{Y} = \{0,1\}^{M + KM}$ to denote the training set labels, where $y_{vu} = 1$ if $e_{vu} \in \mathcal{E}$ and 0 otherwise. 

\textbf{The Cold-Start Problem.} 
The cold-start problem is prevalent in various domains and scenarios. In recommendation systems~\citep{chen2020esam, lu2020meta, hao2021pre, zhu2019addressing, volkovs2017dropoutnet,liu2020long}, cold-start refers to the lack of sufficient interaction history for new users or items, which makes it challenging to provide accurate recommendations. Similarly, in the context of GNNs, the cold-start problem refers to performance in tasks involving cold nodes, which have few or no neighbors in the graph. As illustrated in ~\cref{fig:pre-analysis}, GNNs usually struggle with cold nodes in LP tasks due to unreliable or missing neighbors' information. In this work, we focus on enhancing LP performance for cold nodes, specifically predicting the presence of links between a cold node $v \in \mathcal{V}_{cold}$ and target node $u \in \mathcal{V}$ (w.l.o.g.). Additionally, we aim to maintain satisfactory LP performance for warm nodes. Prior studies on cold-start problems~\citep{tang2020investigating, liu2021tail, zheng2021cold} inspired this research direction.

\section{Node Duplication to Improve Cold-start Performance}
\label{sec:method}
\begin{wrapfigure}{r}{0.6\textwidth}
\vspace{-0.22in}
\begin{algorithm}[H]
\caption{\footnotesize \ours.}\label{alg:main}
\begin{algorithmic}[1]
\footnotesize
\REQUIRE Graph $G = \{\mathcal{V}$, $\mathcal{E}$, $\textbf{X}\}$, Supervision $\mathcal{Y}$, $\text{AGG}$, $\text{UPDATE}$, GNNs Layer $\text{L}$, $\text{DECODER}$, Supervised loss function $\mathcal{L}_{sup}$.
\STATE {\color{red}\textbf{\# Augment the graph by duplicating cold-start nodes $\mathcal{V}_{cold}$.}} 
\STATE Identify cold node set $\mathcal{V}_{cold}$ based on the node degree. 
\STATE \textbf{(Step I)} Duplicate all cold nodes to generate the augmented node set $\mathcal{V}' = \mathcal{V}\cup \mathcal{V}_{cold}$, whose node feature matrix is then $\mathbf{X}' \in \mathbb{R}^{(N+|\mathcal{V}_{cold}|) \times F}$.
\STATE \textbf{(Step II)} Add an edge between each cold node $v \in \mathcal{V}_{cold}$ and its duplication $v'$, then get the augmented edge set $\mathcal{E}' = \mathcal{E}\cup \{e_{vv'}: \forall v \in \mathcal{V}_{cold}\}$.
\STATE \textbf{(Step III)} Add the augmented edges into the training set and get $\mathcal{Y}' = \mathcal{Y} \cup \{y_{vv'}=1: \forall v \in \mathcal{V}_{cold}\}$.
\STATE \textbf{(Step IV)} {\color{blue}\textbf{\# End-to-end supervised training based on the augmented graph $G' = \{\mathcal{V}', \mathcal{E}', \textbf{X}'\}$.}}
\FOR{$l=1$ to $L$}
    \FOR{$v$ in $\mathcal{V'}$}
        \STATE $\boldsymbol{h}'^{(l+1)}_v = \text{UPDATE}\Big(\boldsymbol{h}'^{(l)}_v, \text{AGG}\big(\{\boldsymbol{h}'^{(l)}_u\}: \forall e_{uv} \in \mathcal{E}'\big)\Big)$ 
    \ENDFOR 
\ENDFOR
\FOR{$(i,j)$ in $\mathcal{Y'}$}
    \STATE $\hat{y}'_{ij} = \sigma\big(\text{DECODER}(\boldsymbol{h}'_i, \boldsymbol{h}'_j)\big)$
\ENDFOR
\STATE Loss = $\sum_{(i,j) \in \mathcal{Y}'}\mathcal{L}_{sup}(\hat{y}'_{ij}, y_{ij})$
\end{algorithmic}
\end{algorithm}
\vspace{-0.2in}
\end{wrapfigure}

\vspace{-0.1in}
As described in \cref{sec:preliminary}, a model will not see an isolated node unless it is randomly sampled as a negative training edge for another node in standard supervised LP training. In the same vein, all the cold nodes are strongly underrepresented in the LP training, given their few or even no directly connected neighbors. In light of such observations, our proposed augmentation technique is simple: we duplicate under-represented cold nodes. By both training and aggregating with the edges connecting the cold nodes with their duplications, cold nodes are able to gain better visibility in the training process, which allows the GNN-based LP models to learn better representations. In this section, we introduce \ours in detail, followed by comprehensive analyses of why it works from different perspectives.

\subsection{Proposed Method} 
\label{sec:proposed_method}
We summarize the entire process of \ours in \cref{alg:main}, where the key steps are broken down into four simple stages: \textbf{Step I}: Duplicate all cold nodes to generate the augmented node set $\mathcal{V}' = \mathcal{V}\cup \mathcal{V}_{cold}$, whose node feature matrix is then $\mathbf{X}' \in \mathbb{R}^{(N+|\mathcal{V}_{cold}|) \times F}$; \textbf{Step II}: For each cold node $v \in \mathcal{V}_{cold}$ and its duplication $v'$, add an edge between them and get the augmented edge set $\mathcal{E}' = \mathcal{E}\cup \{e_{vv'}: \forall v \in \mathcal{V}_{cold}\}$; \textbf{Step III}: Include the augmented edges into the training set and get $\mathcal{Y}' = \mathcal{Y} \cup \{y_{vv'}=1: \forall v \in \mathcal{V}_{cold}\}$; \textbf{Step IV}: Proceed with the standard supervised LP training on the augmented graph $G' = \{\mathcal{V}', \mathcal{E}', \textbf{X}'\}$ with augmented training set $\mathcal{Y}'$. Based on extensive experimental analysis, we choose to duplicate code nodes once. The impact of both the type of duplicated nodes and the duplication frequency is further analyzed in~\cref{sec:duplication_timeandnodes}.

\textbf{Time Complexity.} We discuss complexity of our method in terms of the training process on the augmented graph. We use GSage~\citep{hamilton2017inductive} and \innerproduct decoder as the default architecture when demonstrating the following complexity (w.l.o.g). With the augmented graph, GSage has a complexity of $O(R^{L}(N+|\mathcal{V}_{cold}|)D^2)$, where $R$ represents the number of sampled neighbors for each node, $L$ is the number of GSage layers~\citep{wu2020comprehensive}, and $D$ denotes the size of node representations. In comparison to the non-augmented graph, \ours introduces an extra time complexity of $O(R^{L}|\mathcal{V}_{cold}|D^2)$. For the \innerproduct decoder, we incorporate additionally $|\mathcal{V}_{cold}|$ positive edges and also sample $|\mathcal{V}_{cold}|$ negative edges into the training process, resulting in the extra time complexity of the decoder as $O(|\mathcal{V}_{cold}|D)$. Given that all cold nodes have few ($R\le 2$ in our experiments) neighbors, and GSage is also always shallow (so $L$ is small)~\citep{zhao2019pairnorm}, the overall extra complexity introduced by \ours is $O(|\mathcal{V}_{cold}|D^2+|\mathcal{V}_{cold}|D)$.

\subsection{How does Node Duplication Help Cold-start LP?}
\label{sec:analysis}

In this subsection, we analyze how such a simple method can improve cold-start LP from two perspectives: the neighborhood {\bf aggregation} in GNNs and the {\bf supervision} signal during training, in comparison to self-loops in GNNs (e.g., the additional self-connection in the normalized adjacency matrix by GCN). 

\begin{wrapfigure}{r}{0.5\linewidth}
    \centering
    \vspace{-0.15in}
    \includegraphics[width=\linewidth]{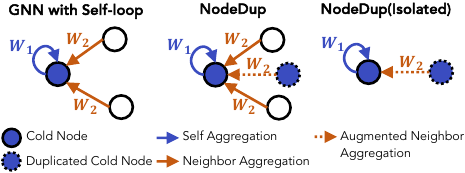}
    \vspace{-0.2in}
    \caption{Comparison of aggregation mechanisms: GNN with Self-loop, \ours for \cold nodes, and \ours for \isolated nodes.}
    \vspace{-0.15in}
    \label{fig:aggregation_figure}
\end{wrapfigure}

\textbf{Aggregation.} As described in \cref{eq:gnn}, when {\small UPDATE}$(\cdot)$ and {\small AGG}$(\cdot)$ do not share the transformation for node features, GNNs would have separate weights for self-representation and neighbor representations, as shown in \cref{fig:aggregation_figure}. The separate weights enable the neighbors and the node itself to play distinct roles in the $\text{\small UPDATE}$ step. By leveraging this property, with \ours, the model can leverage the two ``views'' for each node: first, the existing view is when a node is regarded as the anchor node during message passing, and the additional view is when that node is regarded as one of its neighbors thanks to the duplicated node from \ours. Taking the official PyG~\citep{fey2019fast} implementation of GSage~\citep{hamilton2017inductive} as an example, it updates node representations using $\boldsymbol{h}_v^{(l+1)} = \text{W}_1\boldsymbol{h}_v^{(l)} + \text{W}_2\boldsymbol{m}_{v}^{(l)}$. Here, $\text{W}_1$ and $\text{W}_2$ correspond to the self-representation and neighbors' representations, respectively. Without \ours, isolated nodes $\mathcal{V}_{iso}$ have no neighbors. Thus, the representations of all $v \in \mathcal{V}_{iso}$ are only updated by $\boldsymbol{h}_v^{(l+1)} = \text{W}_1\boldsymbol{h}_v^{(l)}$. With Step II in \ours, the updating process for isolated node $v$ becomes $\boldsymbol{h}_v^{(l+1)} = \text{W}_1\boldsymbol{h}_v^{(l)} + \text{W}_2\boldsymbol{h}_{v}^{(l)} = (\text{W}_1 + \text{W}_2)\boldsymbol{h}_v^{(l)}$. It indicates that $\text{W}_2$ is also incorporated into the node updating process for isolated nodes, which offers an additional perspective for isolated nodes' 
representation learning. Similarly, GAT~\citep{velivckovic2017graph} updates node representations with $\boldsymbol{h}_v^{(l+1)} = \alpha_{vv}\Theta\boldsymbol{h}^{(l)}_v + \sum_{u \in \mathcal{N}_v} \alpha_{vu}\Theta\boldsymbol{h}^{(l)}_u$, where $\alpha_{vu} = \frac{\text{exp}(\text{LeakyReLU}(\boldsymbol{a}^\top[\Theta\boldsymbol{h}^{(l)}_v||\Theta\boldsymbol{h}^{(l)}_u]))}{\sum_{i \in \mathcal{N}_v \cup v}\text{exp}(\text{LeakyReLU}(\boldsymbol{a}^\top[\Theta\boldsymbol{h}^{(l)}_v||\Theta\boldsymbol{h}^{(l)}_i]))}$. Attention scores in $\boldsymbol{a}$ partially correspond to the self-representation $\boldsymbol{h}_v$ and partially to neighbors' representation $\boldsymbol{h}_u$. In this case, neighbor information offers a different perspective compared to self-representation. Such ``multi-view'' enriches the representations learned for the isolated nodes in a similar way to how ensemble methods work~\citep{allen2020towards}. Apart from addressing isolated nodes, the same mechanism and multi-view perspective also apply to \cold nodes.

\begin{wrapfigure}{r}{0.45\linewidth}
    \centering
    \includegraphics[width=\linewidth]{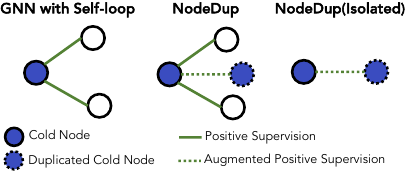}
    \vspace{-0.3in}
    \caption{Comparison of supervision mechanisms.}
    \vspace{-0.2in}
    \label{fig:supervision_figure}
\end{wrapfigure}

\textbf{Supervision.} 
In LP tasks, edges not only facilitate aggregation but also act as positive supervised training signals, as depicted in \cref{fig:supervision_figure}. Cold nodes, which typically have few or no positive training edges, are particularly susceptible to out-of-distribution (OOD) issues~\citep{wu2022handling}, especially in the case of \isolated nodes. Unlike normal self-loops and the self-loops introduced in previous works~\citep{cai2019transgcn, wang2020neighborhood}, where self-loops are solely for aggregation, the edges added by \ours also serve as positive supervision signals for cold nodes through Step III of \cref{alg:main}. By leveraging these additional signals, cold nodes can learn more robust and higher quality embeddings, ultimately improving their performance in LP tasks. 

\begin{wrapfigure}{r}{0.5\linewidth}
    \centering
    \vspace{-0.2in}
    \includegraphics[width=\linewidth]{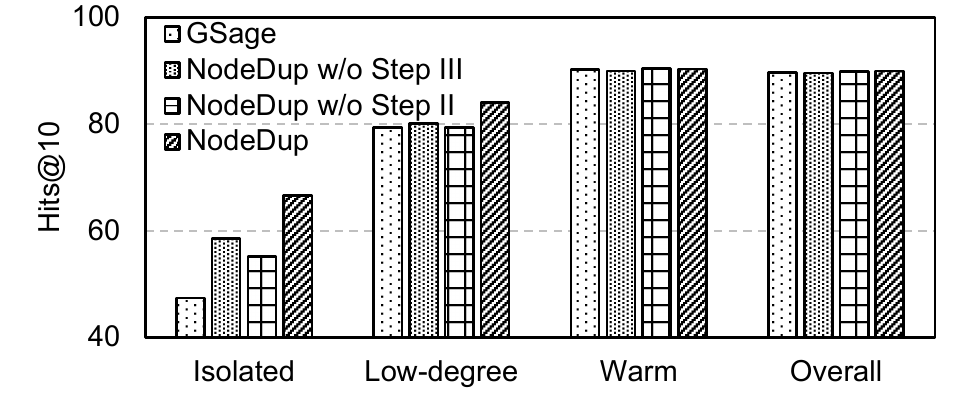}
    \vspace{-0.3in}
    \caption{Ablation study of \ours on \physics. Both Step II and Step III, introduced in \cref{alg:main}, play an important role in performance improvements of \ours.}
    \label{fig:ablation_method}
    \vspace{-0.2in}
\end{wrapfigure}

\textbf{Ablation Study.} \cref{fig:ablation_method} shows an ablation study on these two designs where \ours w/o Step III indicates only using the augmented nodes and edges in aggregation but not supervision; \ours w/o Step II indicates only using the augmented edges in supervision but not aggregation. We can observe that using augmented nodes and edges either in supervision or aggregation can significantly improve the LP performance on \isolated nodes. By combining them, \ours results in larger improvements. Besides, \ours also achieves improvements on \cold nodes while not sacrificing the performance on \warm nodes.

\begin{figure*}
    \centering
    \includegraphics[width=\linewidth]{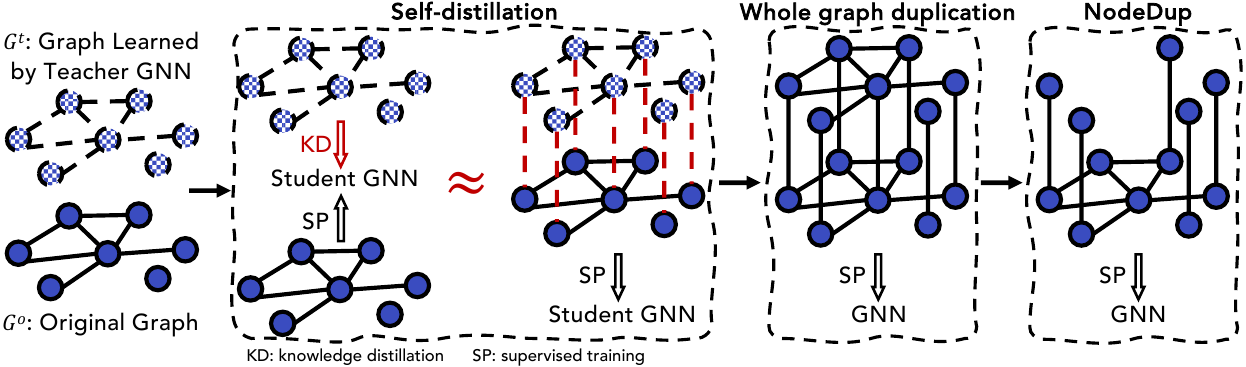}
    \caption{Comparing \ours to self-distillation. The self-distillation process can be approximated by training the student GNN on an augmented graph, which combines $G^o$, $G^t$, and edges connecting corresponding nodes in the two graphs. This process can be further improved by replacing $G^t$ with $G^o$ to explore the whole graph duplication. \ours is a lightweight variation of it.}
    \label{fig:KD}
\end{figure*}

\subsection{Further Insight: Understanding \ours through Self-distillation} 

As introduced in \cref{sec:analysis}, the effectiveness of \ours arises from leveraging diverse perspectives or signals obtained during both the aggregation and supervision steps. \citet{allen2020towards} showed that the success of self-distillation, similar to our method, contributes to ensemble learning by providing models with different perspectives on the knowledge. Building on this insight, we show an interesting interpretation of \ours, positioning it as a simplified and enhanced adaptation of self-distillation for LP tasks for cold nodes, as illustrated in \cref{fig:KD}, in which we draw a connection between self-distillation and \ours.

\begin{wrapfigure}{r}{0.5\linewidth}
    \centering
    \includegraphics[width=\linewidth]{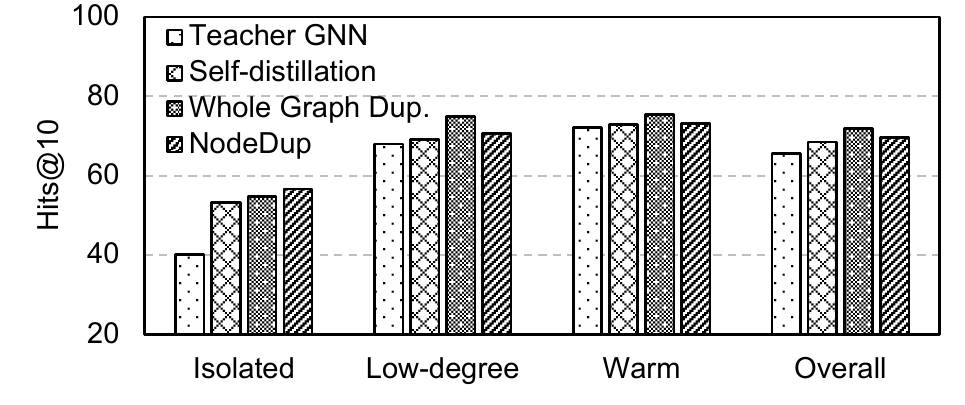}
    \vspace{-0.3in}
    \caption{Performance with different training strategies introduced in \cref{fig:KD} on \citeseer. \ours achieves better performance across all settings.}
    \label{fig:KD_results}
    \vspace{-0.2in}
\end{wrapfigure}

In \textbf{self-distillation}, a teacher GNN is first trained to learn the node representations $\mathbf{H}^t$ from original features $\mathbf{X}$ through supervised training for LP tasks on the original graph. We denote the original graph as $G^o$, and we denote the graph, where we replace the node features in $G^o$ with $\mathbf{H}^t$, as $G^t$ in~\cref{fig:KD}. The student GNN is then initialized with random parameters and trained with the sum of two loss functions: $\mathcal{L}_{SD} = \mathcal{L}_{SP}+\mathcal{L}_{KD}$, where $\mathcal{L}_{SP}$ denotes the supervised training loss with $G^o$ and $\mathcal{L}_{KD}$ denotes the knowledge distillation loss with $G^t$. \cref{fig:KD_results} shows that self-distillation outperforms the teacher GNN across all settings. 

The effect of $\mathcal{L}_{KD}$ is similar to that of creating an additional link connecting nodes in $G^o$ to their corresponding nodes in $G^t$ when optimizing with $\mathcal{L}_{SP}$. This is illustrated by the red dashed line in~\cref{fig:KD}. For better clarity, we show the similarities between these two when we use the \innerproduct as the decoder for LP with the following example. Given a node $v$ with normalized teacher embedding $\boldsymbol{h}_v^t$ and normalized student embedding $\boldsymbol{h}_v$, the additional loss term that would be added for distillation with cosine similarity is $\mathcal{L}_{KD} = - \frac{1}{N} \sum_{v \in \mathcal{V}}  \boldsymbol{h}_v \cdot \boldsymbol{h}_v^t$. On the other hand, for the dashed line edges in~\cref{fig:KD}, we add an edge between the node $v$ and its corresponding node $v'$ in $G^t$ with embedding $\boldsymbol{h}_{v'}^t$. When trained with an \innerproduct decoder and binary cross-entropy loss, it results in the following: $\mathcal{L}_{SP} =  - \frac{1}{N} \sum y_{vv'} \log( \boldsymbol{h}_v \cdot \boldsymbol{h}_{v'}^t ) + (1-y_{vv'}) \log (1 - \boldsymbol{h}_v \cdot \boldsymbol{h}_{v'}^t)$. Since we always add the edge $(v, v')$, we know $y_{vv'} = 1$, and can simplify the loss as follows: $\mathcal{L}_{SP} = - \frac{1}{N} \sum \log( \boldsymbol{h}_v \cdot \boldsymbol{h}_{v'}^t )$. Here, we can observe that $\mathcal{L}_{KD}$ and $\mathcal{L}_{SP}$ are positively correlated as $\log(\cdot)$ is a monotonically increasing function.

To further improve this step and mitigate potential noise in $G^t$, we explore a whole graph duplication technique, where $G^t$ is replaced with an exact duplicate of $G^o$ to train the student GNN. The results in \cref{fig:KD_results} demonstrate significant performance enhancement achieved by \textbf{whole graph duplication} compared to self-distillation. \ours is a lightweight variation of the whole graph duplication technique, which focuses on duplicating only the cold nodes and adding edges connecting them to their duplicates. From the results, it is evident that \textbf{\ours} consistently outperforms the teacher GNN and self-distillation in all scenarios. Additionally, \ours exhibits superior performance on isolated nodes and is much more efficient compared to the whole graph duplication approach.

\subsection{\ourslight: An Efficient Variant of \ours}

\begin{wrapfigure}{r}{0.35\linewidth}
    \centering
    \vspace{-0.2in}
    \includegraphics[width=\linewidth]{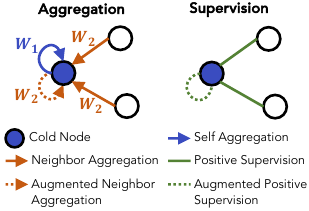}
    \vspace{-0.35in}
    \caption{Aggregation and supervision mechanisms of \ourslight.}
    \label{fig:nodedupL_figure}
    \vspace{-0.25in}
\end{wrapfigure}

Inspired by the above analysis, we further introduce a lightweight variant of \ours for better efficiency, \ourslight. To provide above-described ``multi-view'' information as well as the supervision signals for cold nodes, \ourslight simply add additional self-loop edges for the cold nodes into the edge set $\mathcal{E}$, that is, $\mathcal{E}' = \mathcal{E} \cup \{e_{vv}: \forall v \in \mathcal{V}_{cold}\}$. During aggregation, \ourslight intentionally incorporates the self-representation $\boldsymbol{h}_v^{(l)}$ into the aggregated neighbors' representation $\boldsymbol{m}^{(l)}_v$ through these additional edges. This allows the weight matrix $\text{W}_2$ to provide an extra ``view'' of $\boldsymbol{h}_v^{(l)}$ when updating $\boldsymbol{h}_v^{(l+1)}$. For supervision, the added edges also serve as positive training samples for cold nodes. As demonstrated in \cref{fig:nodedupL_figure}, \ourslight preserves the two essential designs of \ours while avoiding the addition of extra nodes, which further saves time and space complexity. Moreover, \ours differs from \ourslight since each duplicated node in \ours will provide another view for itself because of dropout layers, which leads to different performance as shown in~\cref{sec:base}.

\section{Experiments}
\subsection{Experimental Settings}
\label{sec:experimental_settings}
\textbf{Datasets and Evaluation Settings.} We conduct experiments on 7 benchmark datasets: \cora, \citeseer, \cs, \physics, \computers, \photos and \igbtiny, with their details specified in \cref{sec:data}. We randomly split edges into training, validation, and testing sets. We allocated 10\% for validation and 40\% for testing in \computers and \photos, 5\%/10\% for testing in \igbtiny, and 10\%/20\% in other datasets. We follow the standard evaluation metrics used in the Open Graph Benchmark~\citep{hu2020open} for LP, in which we rank missing references higher than 500 negative reference candidates for each node. The negative references are randomly sampled from nodes not connected to the source node. We use Hits@10 as the main evaluation metric~\citep{han2022mlpinit}. We follow \citet{guo2022linkless} and \citet{shiao2022link} for the inductive settings, where new nodes appear after the training process. Additionally, results for large-scale datasets and heterophilic graphs are presented in~\cref{sec:largescale} and~\cref{sec:heterophily}.

\textbf{Baselines.} Both \ours and \ourslight are flexible to integrate with different GNN encoder architectures and LP decoders. For our experiments, we use GSage~\citep{hamilton2017inductive} encoder and the \innerproduct decoder as the default base LP model. To comprehensively evaluate our work, we compare \ours against three categories of baselines. (1) Base LP models. (2) Cold-start methods: TailGNN~\citep{liu2021tail} and Cold-brew~\citep{zheng2021cold} primarily aim to enhance the performance on cold nodes. We also compared with Imbalance~\citep{lin2017focal}, viewing cold nodes as an issue of the imbalance concerning node degrees. (3) Graph data augmentation methods: Augmentation frameworks including DropEdge~\citep{rong2019dropedge}, TuneUP~\citep{hu2022tuneup}, and LAGNN~\citep{liu2022local} typically improve the performance while introducing additional preprocessing or training time. Performance comparisons with heuristic methods and additional cold-start methods (e.g. Upsampling, DegFairGNN~\citep{liu2023generalized}, SAILOR~\citep{liao2023sailor} and GRADE~\citep{wang2022uncovering}) are in~\cref{sec:heuristic} and \cref{sec:appendix-cold}.

\begin{table}[t]
\centering
\small
\vspace{-0.1in}
\caption{Performance compared with base GNN and baselines for cold-start methods. The best result is \textbf{bold}, and the runner-up is \underline{underlined}. \ours and \ourslight outperform GSage and cold-start baselines almost all the cases. }
\label{tab:tailnodes}
\scalebox{0.9}{
\begin{tabular}{l|c||c|ccc||cc} 
\toprule
\multicolumn{2}{c||}{}&GSage &Imbalance &TailGNN &Cold-brew &\ourslight &\ours\\ \midrule
\multirow{4}{*}{\cora}&\isolated & \ms{32.20}{3.58} & \ms{34.51}{1.11} & \ms{36.95}{1.34} & \ms{28.17}{0.67} & \ms{\underline{39.76}}{1.32} & \ms{\textbf{44.27}}{3.82} \\
&\cold &\ms{59.45}{1.09} & \ms{59.42}{1.21} & \ms{61.35}{0.79} & \ms{57.27}{0.63} & \ms{\textbf{62.53}}{1.03} & \ms{\underline{61.98}}{1.14} \\
&\warm &\ms{\underline{61.14}}{0.78} & \ms{59.54}{0.46} & \ms{60.61}{0.90} & \ms{56.28}{0.81} & \ms{\textbf{62.07}}{0.37} & \ms{59.07}{0.68} \\
&\overall &\ms{58.31}{0.68} & \ms{57.55}{0.67} & \ms{\underline{59.02}}{0.71} & \ms{54.44}{0.53} & \ms{\textbf{60.49}}{0.49} & \ms{58.92}{0.82}\\\midrule
\multirow{4}{*}{\citeseer}&\isolated &\ms{47.13}{2.43} & \ms{46.26}{0.86} & \ms{37.84}{3.36} & \ms{37.78}{4.23} & \ms{\underline{52.46}}{1.16} & \ms{\textbf{57.54}}{1.04} \\
&\cold &\ms{61.88}{0.79} & \ms{61.90}{0.60} & \ms{62.06}{1.73} & \ms{59.12}{9.97} & \ms{\underline{73.71}}{1.22} & \ms{\textbf{75.50}}{0.39} \\
&\warm &\ms{71.45}{0.52} & \ms{71.54}{0.86} & \ms{71.32}{1.83} & \ms{65.12}{7.82} & \ms{\textbf{74.99}}{0.37} & \ms{\underline{74.68}}{0.67} \\
&\overall &\ms{63.77}{0.83} & \ms{63.66}{0.43} & \ms{62.02}{1.89} & \ms{58.03}{7.72} & \ms{\underline{70.34}}{0.35} & \ms{\textbf{71.73}}{0.47}\\\midrule
\multirow{4}{*}{\cs}&\isolated &\ms{56.41}{1.61} & \ms{46.60}{1.66} & \ms{55.70}{1.38} & \ms{57.70}{0.81} & \ms{\underline{65.18}}{1.25} & \ms{\textbf{65.87}}{1.70} \\
&\cold &\ms{75.95}{0.25} & \ms{75.53}{0.21} & \ms{73.60}{0.70} & \ms{73.99}{0.34} & \ms{\textbf{81.46}}{0.57} & \ms{\underline{81.12}}{0.36} \\
&\warm &\ms{84.37}{0.46} & \ms{83.70}{0.46} & \ms{79.86}{0.35} & \ms{78.23}{0.28} & \ms{\textbf{85.48}}{0.26} & \ms{\underline{84.76}}{0.41} \\
&\overall &\ms{83.33}{0.42} & \ms{82.56}{0.40} & \ms{79.05}{0.36} & \ms{77.63}{0.23} & \ms{\textbf{84.90}}{0.29} & \ms{\underline{84.23}}{0.39}\\\midrule
\multirow{4}{*}{\physics}&\isolated &\ms{47.41}{1.38} & \ms{55.01}{0.58} & \ms{52.54}{1.34} & \ms{64.38}{0.85} & \ms{\underline{65.04}}{0.63} & \ms{\textbf{66.65}}{0.95} \\
&\cold &\ms{79.31}{0.28} & \ms{79.50}{0.27} & \ms{75.95}{0.27} & \ms{75.86}{0.10} & \ms{\underline{82.70}}{0.22} & \ms{\textbf{84.04}}{0.22} \\
&\warm &\ms{90.28}{0.23} & \ms{89.85}{0.09} & \ms{85.93}{0.40} & \ms{78.48}{0.14} & \ms{\textbf{90.44}}{0.23} & \ms{\underline{90.33}}{0.05} \\
&\overall &\ms{89.76}{0.22} & \ms{89.38}{0.09} & \ms{85.48}{0.38} & \ms{78.34}{0.13} & \ms{\textbf{90.09}}{0.22} & \ms{\underline{90.03}}{0.05}\\\midrule
\multirow{4}{*}{\computers}&\isolated &\ms{9.32}{1.44 } & \ms{10.14}{0.59} & \ms{10.63}{1.59} & \ms{9.75}{1.24 } & \ms{\underline{17.11}}{1.62} & \ms{\textbf{19.62}}{2.63} \\
&\cold &\ms{57.91}{0.97} & \ms{56.19}{0.82} & \ms{51.21}{1.58} & \ms{49.03}{0.94} & \ms{\textbf{62.14}}{1.06} & \ms{\underline{61.16}}{0.92} \\
&\warm &\ms{66.87}{0.47} & \ms{65.62}{0.21} & \ms{62.77}{0.44} & \ms{57.52}{0.28} & \ms{\underline{68.02}}{0.41} & \ms{\textbf{68.10}}{0.25} \\
&\overall &\ms{66.67}{0.47} & \ms{65.42}{0.20} & \ms{62.55}{0.45} & \ms{57.35}{0.28} & \ms{\underline{67.86}}{0.41} & \ms{\textbf{67.94}}{0.25}\\\midrule
\multirow{4}{*}{\photos}&\isolated &\ms{9.25}{2.31 } & \ms{10.80}{1.72} & \ms{13.62}{1.00} & \ms{12.86}{2.58} & \ms{\textbf{21.50}}{2.14} & \ms{\underline{17.84}}{3.53} \\
&\cold &\ms{52.61}{0.88} & \ms{50.68}{0.57} & \ms{42.75}{2.50} & \ms{43.14}{0.64} & \ms{\textbf{55.70}}{1.38} & \ms{\underline{54.13}}{1.58} \\
&\warm &\ms{67.64}{0.55} & \ms{64.54}{0.50} & \ms{61.63}{0.73} & \ms{58.06}{0.56} & \ms{\textbf{69.68}}{0.87} & \ms{\underline{68.68}}{0.49} \\
&\overall &\ms{67.32}{0.54} & \ms{64.24}{0.49} & \ms{61.29}{0.75} & \ms{57.77}{0.56} & \ms{\textbf{69.40}}{0.86} & \ms{\underline{68.39}}{0.48}\\\midrule
\multirow{4}{*}{\igbtiny}&\isolated &\ms{75.92}{0.52} & \ms{77.32}{0.79} & \ms{77.29}{0.34} & \ms{82.31}{0.30} & \ms{\underline{87.43}}{0.44} & \ms{\textbf{88.04}}{0.20} \\
&\cold &\ms{79.38}{0.23} & \ms{79.19}{0.09} & \ms{80.57}{0.14} & \ms{83.84}{0.16} & \ms{\underline{88.37}}{0.24} & \ms{\textbf{88.98}}{0.17} \\
&\warm &\ms{86.42}{0.24} & \ms{86.01}{0.19} & \ms{85.35}{0.19} & \ms{82.44}{0.21} & \ms{\textbf{88.54}}{0.31} & \ms{\underline{88.28}}{0.20} \\
&\overall &\ms{84.77}{0.21} & \ms{84.47}{0.14} & \ms{84.19}{0.18} & \ms{82.68}{0.17} & \ms{\textbf{88.47}}{0.28} & \ms{\underline{88.39}}{0.18}\\
\bottomrule
\end{tabular}}
\vspace{-0.1in}
\end{table}

\subsection{Performance Compared to Base GNN LP Models}
\label{sec:base}
\textbf{\isolated and \cold Nodes. }We compare our methods with base GNN LP models that consist of a GNN encoder in conjunction with an \innerproduct decoder and are trained with a supervised loss. From \cref{tab:tailnodes}, we observe consistent improvements for both \ourslight and \ours over the base GSage model across all datasets, particularly in the \isolated and \cold node settings. 
Notably, in the \isolated setting, \ours achieves an impressive 29.6\% improvement, on average, across all datasets. These findings provide clear evidence that our methods effectively address the issue of sub-optimal LP performance on cold nodes.

\noindent \textbf{Warm Nodes and Overall.} It is encouraging to see that \ourslight consistently outperforms GSage across all the datasets in the \warm nodes and \overall settings. \ours also outperforms GSage in 13 out of 14 cases under both settings. These findings support the notion that our methods can effectively maintain and enhance the performance of \warm nodes. \textbf{Why?} The superior performance on \warm nodes is directly tied to our focus on LP tasks, where we evaluate node pair outcomes. Given the substantial number of Warm-Cold node pairs under prediction, these outcomes contribute to the overall performance metrics for both Warm node prediction. Better learning of Cold nodes thus boosts Cold-Warm node pairs link prediction performance, which subsequently elevates the prediction accuracy for \warm nodes. A more detailed experimental analysis is provided in~\cref{appendix:warm-warm analysis}.

\noindent \textbf{\ours vs. \ourslight.} Furthermore, we observe that \ours achieves greater improvements over \ourslight for \isolated nodes. However, \ourslight outperforms \ours on 6 out of 7 datasets for \warm nodes. The additional improvements achieved by \ours for \isolated nodes can be attributed to the extra view provided to cold nodes through node duplication during aggregation. On the other hand, the impact of node duplication on the original graph structure likely affects the performance of \warm nodes, which explains the superior performance of \ourslight in this setting compared to \ours. 

\begin{figure}[h]
    \centering
    \includegraphics[width=0.9\linewidth]{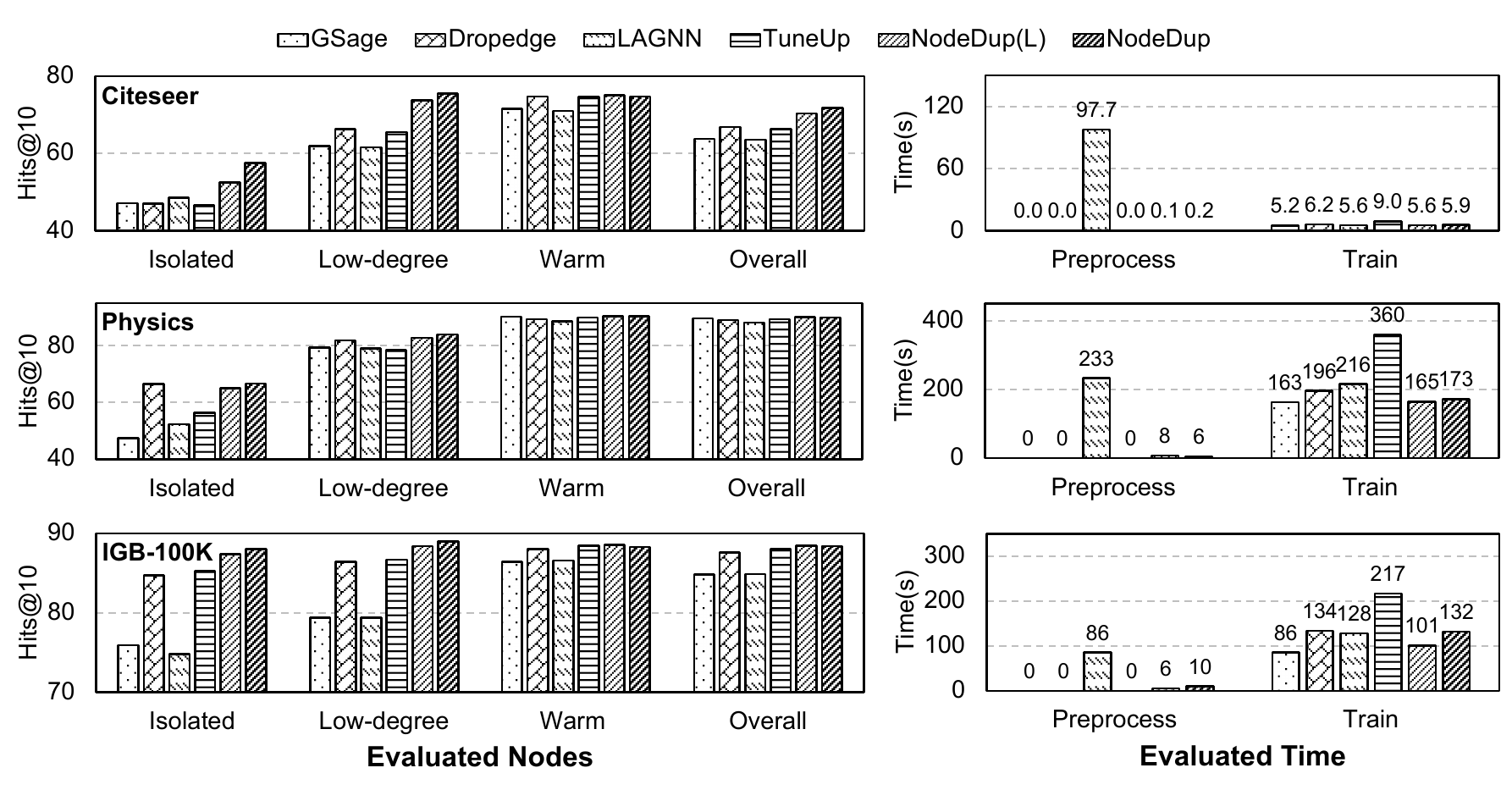}
    \vspace{-0.2in}
    \caption{Performance and runtime comparisons of different augmentation methods. The \textit{left} histograms show the performance results, and the \textit{right} histograms show the preprocessing and training time consumption of each method. Our methods consistently achieve significant improvements in both performance for \isolated and \cold node settings and runtime efficiency over baselines.}
    \vspace{-0.2in}
    \label{fig:aug}
\end{figure}

\subsection{Performance Compare to Cold-start Methods}
\cref{tab:tailnodes} presents the LP performance of various cold-start baselines. For both \isolated and \cold nodes, we consistently observe \textit{substantial improvements} of our \ours and \ourslight methods compared to other cold-start baselines. Specifically, \ours and \ourslight achieve 38.49\% and 34.74\% improvement for \isolated nodes on average across all datasets, respectively. 

In addition, our methods consistently outperform cold-start baselines for \warm nodes across all the datasets, where \ourslight and \ours achieve 6.76\% and 7.95\% improvements on average, respectively. 
This shows that our methods can successfully overcome issues with degrading performance on \warm nodes in cold-start baselines. Further analyses with other cold-start methods and efficiency comparisons can be found in~\cref{sec:appendix-cold} and~\cref{sec:appendix_coldtime}.

\vspace{-0.1in}
\subsection{Performance Compared to Augmentation Methods}
\label{sec:augment}
\textbf{Effectiveness Comparison.} Since \ours and \ourslight use graph data augmentation techniques, we compare them to other data augmentation baselines. 
The performance and time consumption results for \citeseer, \physics, and \igbtiny are shown in \cref{fig:aug}, with additional datasets in \cref{sec:append_aug} due to space constraints.
From \cref{fig:aug}, \ours consistently outperforms all the graph augmentation baselines for \isolated and \cold nodes across all three datasets, while \ourslight outperforms baselines in 17/18 cases for \isolated and \cold nodes.
Both \ours and \ourslight also perform on par or above baselines for \warm nodes.

\textbf{Efficiency Comparison.} Augmentation methods often come with the trade-off of adding additional run time before or during model training. For example, LAGNN~\citep{liu2022local} requires extra preprocessing time to train the generative model prior to GNN training. It also takes additional time to generate extra features for each node during training. Although Dropedge~\citep{rong2019dropedge} and TuneUP~\citep{hu2022tuneup} are free of preprocessing, they require additional time to drop edges in each training epoch compared to base GNN training. Furthermore, the two-stage training employed by TuneUP doubles the training time compared to one-stage training methods. For \ours methods, duplicating nodes and adding edges is remarkably swift and consumes significantly less preprocessing time than other augmentation methods. As an example, \ourslight and \ours are \textbf{977.0$\times$} and \textbf{488.5$\times$} faster than LAGNN in preprocessing \citeseer, respectively. We also observe that \ourslight has the least training time among all augmentation methods and datasets, while \ours also requires less training time in 8/9 cases. Additionally, \ourslight achieves significant efficiency benefits compared to \ours in \cref{fig:aug}, especially when the number of nodes in the graph increases substantially. Taking the \igbtiny dataset as an example, \ourslight is 1.3$\times$ faster than \ours for the entire training process.

\begin{wraptable}{r}{0.5\linewidth}
\vspace{-0.3in}
\caption{Performance in inductive settings. The best result is \textbf{bold}, and the runner-up is \underline{underlined}. Our methods consistently outperform GSage.}
\label{tab:product}
\scalebox{0.76}{
\begin{tabular}{l|c|ccc} 
\toprule
\multicolumn{2}{c|}{}&GSage &\ourslight &\ours \\\midrule
\multirow{4}{*}{\citeseer}&\isolated &\ms{58.42}{0.49} & \ms{\underline{62.42}}{1.88} & \ms{\textbf{62.94}}{1.91} \\
&\cold &\ms{67.75}{1.06} & \ms{\underline{69.93}}{1.18} & \ms{\textbf{72.05}}{1.23} \\
&\warm &\ms{72.98}{1.15} & \ms{\textbf{75.04}}{1.03} & \ms{\underline{74.40}}{2.43} \\
&\overall &\ms{66.98}{0.61} & \ms{\underline{69.65}}{0.83} & \ms{\textbf{70.26}}{1.16}\\\midrule
\multirow{4}{*}{\physics}&\isolated &\ms{85.62}{0.23} & \ms{\underline{85.94}}{0.15} & \ms{\textbf{86.90}}{0.35} \\
&\cold &\ms{80.87}{0.43} & \ms{\underline{81.23}}{0.56} & \ms{\textbf{85.56}}{0.25} \\
&\warm &\ms{90.22}{0.36} & \ms{\underline{90.37}}{0.25} & \ms{\textbf{90.54}}{0.14} \\
&\overall &\ms{89.40}{0.33} & \ms{\underline{89.57}}{0.23} & \ms{\textbf{89.98}}{0.13}\\\midrule
\multirow{4}{*}{\igbtiny}&\isolated &\ms{84.33}{0.87} & \ms{\underline{92.94}}{0.11} & \ms{\textbf{93.95}}{0.06} \\
&\cold &\ms{93.19}{0.06} & \ms{\underline{93.33}}{0.11} & \ms{\textbf{94.00}}{0.09} \\
&\warm &\ms{90.76}{0.13} & \ms{\textbf{91.21}}{0.07} & \ms{\underline{91.20}}{0.08} \\
&\overall &\ms{90.31}{0.18} & \ms{\underline{91.92}}{0.05} & \ms{\textbf{92.21}}{0.04}\\
\bottomrule
\end{tabular}}
\vspace{-0.2in}
\end{wraptable}

\vspace{-0.1in}
\subsection{Performance under the Inductive Setting}
Under the inductive setting \citep{guo2022linkless, shiao2022link},  which closely resembles real-world LP scenarios, the presence of new nodes after the training stage adds an additional challenge compared to the transductive setting. We evaluate and present the effectiveness of our methods under this setting in \cref{tab:product} for \citeseer, \physics, and \igbtiny datasets. Additional results for other datasets can be found in \cref{sec:append_induc}. In \cref{tab:product}, we observe that our methods consistently outperform base GSage across all of the datasets. We also observe significant performance improvements of our methods on \isolated nodes, where \ours and \ourslight achieve 5.50\% and 3.57\% improvements averaged across the three datasets, respectively. Additionally, \ours achieves 5.09\% improvements on \cold nodes. \ours leads to more pronounced improvements on \cold/\isolated nodes, making it particularly beneficial for the inductive setting. 

\subsection{Performance with Different Encoders/Decoders}
As a simple plug-and-play augmentation method, \ours can work with different GNN encoders and LP decoders. In \cref{tab:ablation_encoder,tab:ablation_decoder}, we present results with GAT~\citep{velivckovic2017graph} and JKNet~\citep{xu2018representation} as encoders, along with a MLP decoder. Due to the space limit, we only report the results of three datasets here and leave the remaining in \cref{append:ablation_study}. When applying \ours to base LP training, with GAT or JKNet as the encoder and \innerproduct as the decoder, we observe significant performance improvements across the board. Regardless of the encoder choice, \ours consistently outperforms the base models, particularly for \isolated and \cold nodes. From \cref{append:ablation_study}, we also observe the performance improvements of \ours with GCN~\citep{kipf2016semi}, GraphTransformer~\citep{dwivedi2020generalization} as encoders.

\begin{table}
\centering
\caption{Performance with different encoders (inner product as the decoder). The best result for each encoder is \textbf{bold}, and the runner-up is \underline{underlined}. Our methods consistently outperform the base models, particularly for \isolated and \cold nodes.}
\label{tab:ablation_encoder}
\scalebox{0.8}{
\begin{tabular}{l|c||ccc||ccc} 
\toprule
\multicolumn{2}{c||}{}&GAT &\ourslight &\ours &JKNet &\ourslight &\ours\\ \midrule
\multirow{4}{*}{\citeseer}&\isolated&\ms{37.78}{2.36} & \ms{\underline{38.95}}{2.75} & \ms{\textbf{44.04}}{1.03} & \ms{37.78}{0.63} & \ms{\underline{49.06}}{0.60} & \ms{\textbf{55.15}}{0.87} \\
&\cold &\ms{58.04}{2.40} & \ms{\underline{61.93}}{1.66} & \ms{\textbf{66.73}}{0.96} & \ms{60.74}{1.18} & \ms{\underline{71.78}}{0.64} & \ms{\textbf{75.26}}{1.16} \\
&\warm &\ms{56.37}{2.15} & \ms{\underline{64.55}}{1.74} & \ms{\textbf{66.61}}{1.67} & \ms{71.61}{0.76} & \ms{\underline{74.66}}{0.47} & \ms{\textbf{75.81}}{0.89} \\
&\overall &\ms{53.42}{1.59} & \ms{\underline{58.89}}{0.89} & \ms{\textbf{62.41}}{0.78} & \ms{61.73}{0.57} & \ms{\underline{68.91}}{0.38} & \ms{\textbf{71.75}}{0.82}\\\midrule
\multirow{4}{*}{\physics}&\isolated&\ms{38.19}{1.23} & \ms{\underline{39.95}}{1.48} & \ms{\textbf{45.89}}{2.82} & \ms{42.57}{1.93} & \ms{\underline{55.47}}{2.25} & \ms{\textbf{61.11}}{2.27} \\
&\cold &\ms{74.19}{0.31} & \ms{\underline{74.77}}{0.46} & \ms{\textbf{76.36}}{0.25} & \ms{75.36}{0.23} & \ms{\underline{79.55}}{0.21} & \ms{\textbf{81.14}}{0.28} \\
&\warm &\ms{85.84}{0.32} & \ms{\textbf{86.02}}{0.45} & \ms{\underline{85.84}}{0.15} & \ms{88.24}{0.32} & \ms{\textbf{89.42}}{0.16} & \ms{\underline{89.24}}{0.16} \\
&\overall &\ms{85.27}{0.30} & \ms{\textbf{85.47}}{0.45} & \ms{\underline{85.37}}{0.14} & \ms{87.64}{0.31} & \ms{\textbf{88.96}}{0.15} & \ms{\underline{88.87}}{0.15}\\\midrule
\multirow{4}{*}{\igbtiny}&\isolated&\ms{75.87}{0.48} & \ms{\underline{78.17}}{0.58} & \ms{\textbf{80.18}}{0.31} & \ms{69.29}{0.73} & \ms{\underline{86.60}}{0.46} & \ms{\textbf{86.85}}{0.41} \\
&\cold &\ms{77.05}{0.15} & \ms{\underline{78.50}}{0.31} & \ms{\textbf{81.00}}{0.12} & \ms{76.90}{0.27} & \ms{\underline{86.94}}{0.15} & \ms{\textbf{87.65}}{0.20} \\
&\warm &\ms{\underline{81.40}}{0.07} & \ms{\textbf{81.95}}{0.25} & \ms{81.19}{0.20} & \ms{84.93}{0.30} & \ms{\textbf{87.41}}{0.13} & \ms{\underline{86.19}}{0.12} \\
&\overall &\ms{80.42}{0.07} & \ms{\textbf{81.19}}{0.25} & \ms{\underline{81.11}}{0.19} & \ms{82.91}{0.28} & \ms{\textbf{87.29}}{0.13} & \ms{\underline{86.47}}{0.13}\\
\bottomrule
\end{tabular}}
\vspace{-0.1in}
\end{table}

\begin{wraptable}{r}{0.5\linewidth}
\vspace{-0.3in}
\caption{LP performance with MLP decoder (GSage as the encoder). Our methods outperform the base model.}
\label{tab:ablation_decoder}
\scalebox{0.75}{
\begin{tabular}{l|c||ccc} 
\toprule
\multicolumn{2}{c||}{}&MLP-Dec. &\ourslight &\ours \\ \midrule
\multirow{4}{*}{\citeseer}&\isolated&\ms{17.16}{1.14} & \ms{\underline{37.84}}{3.06} & \ms{\textbf{51.17}}{2.19} \\
&\cold &\ms{63.82}{1.58} & \ms{\underline{68.49}}{1.19} & \ms{\textbf{71.98}}{1.29} \\
&\warm &\ms{72.93}{1.25} & \ms{\underline{75.33}}{0.54} & \ms{\textbf{75.72}}{0.55} \\
&\overall &\ms{59.49}{1.21} & \ms{\underline{66.07}}{0.74} & \ms{\textbf{69.89}}{0.65}\\\midrule
\multirow{4}{*}{\physics}&\isolated&\ms{11.59}{1.88} & \ms{\textbf{60.25}}{2.54} & \ms{\underline{59.50}}{1.87} \\
&Low-degree &\ms{76.37}{0.64} & \ms{\underline{81.74}}{0.77} & \ms{\textbf{82.58}}{0.79} \\
&\warm &\ms{91.54}{0.33} & \ms{\textbf{91.96}}{0.36} & \ms{\underline{91.59}}{0.22} \\
&\overall &\ms{90.78}{0.33} & \ms{\textbf{91.51}}{0.38} & \ms{\underline{91.13}}{0.23}\\\midrule
\multirow{4}{*}{\igbtiny}&\isolated&\ms{3.51}{0.32 } & \ms{\textbf{82.71}}{1.05} & \ms{\underline{82.02}}{0.73} \\
&\cold &\ms{75.25}{0.49} & \ms{\underline{85.96}}{0.42} & \ms{\textbf{86.04}}{0.26} \\
&\warm &\ms{85.06}{0.08} & \ms{\textbf{87.89}}{0.13} & \ms{\underline{86.87}}{0.48} \\
&\overall &\ms{80.16}{0.16} & \ms{\textbf{87.35}}{0.21} & \ms{\underline{86.54}}{0.40}\\
\bottomrule
\end{tabular}}
\vspace{-0.2in}
\end{wraptable}

In \cref{tab:ablation_decoder}, we present the results of our methods applied to the base LP training, where GSage serves as the encoder and MLP as the decoder. 
Regardless of the decoder, we observe better performance with our methods. These improvements are significantly higher compared to the improvements observed with the \innerproduct decoder. The primary reason for this discrepancy is the inclusion of additional supervised training signals for isolated nodes in our methods, as discussed in \cref{sec:analysis}. These signals play a crucial role in training the MLP decoder, making it more responsive to the specific challenges presented by isolated nodes. 

Furthermore, we apply \ours to SEAL~\citep{zhang2018link}, a subgraph-based LP model, and observe notable performance gains, as shown in \cref{append:ablation_study}, demonstrating the broad applicability of our method across different GNN models and training paradigms.

\subsection{Influence of the Duplication Frequency and Duplicated Node Types in \ours}
\label{sec:duplication_timeandnodes}

\begin{figure}[t]
    \centering
    \vspace{-0.2in}
    \subfigure[\isolated]
    {\includegraphics[width=0.32\linewidth]{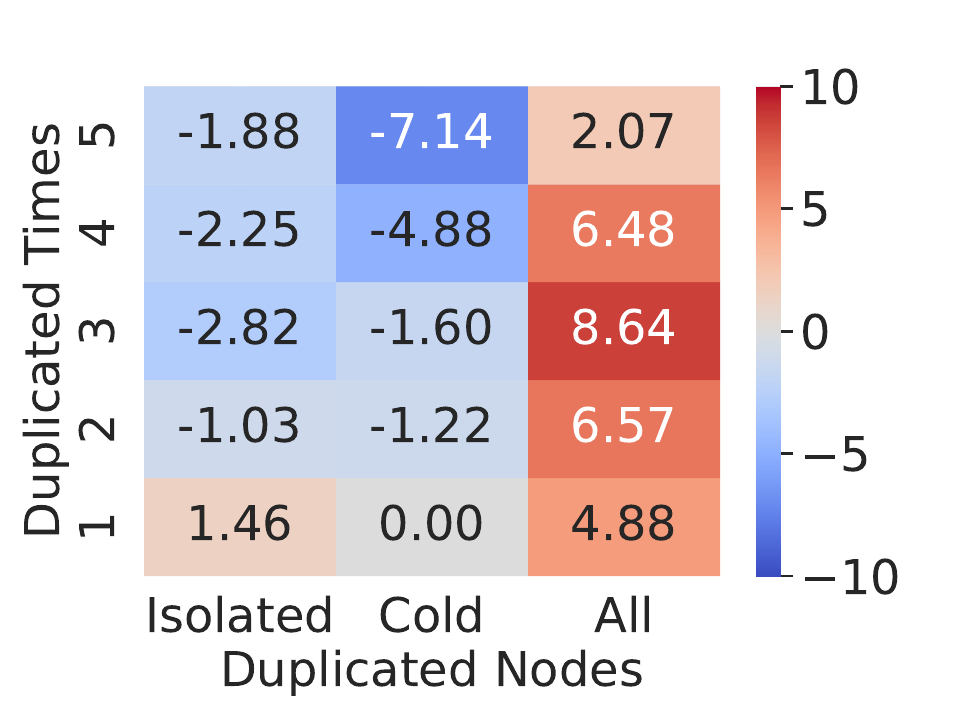}\label{fig:isolated}}
    \subfigure[\cold]
    {\includegraphics[width=0.32\linewidth]{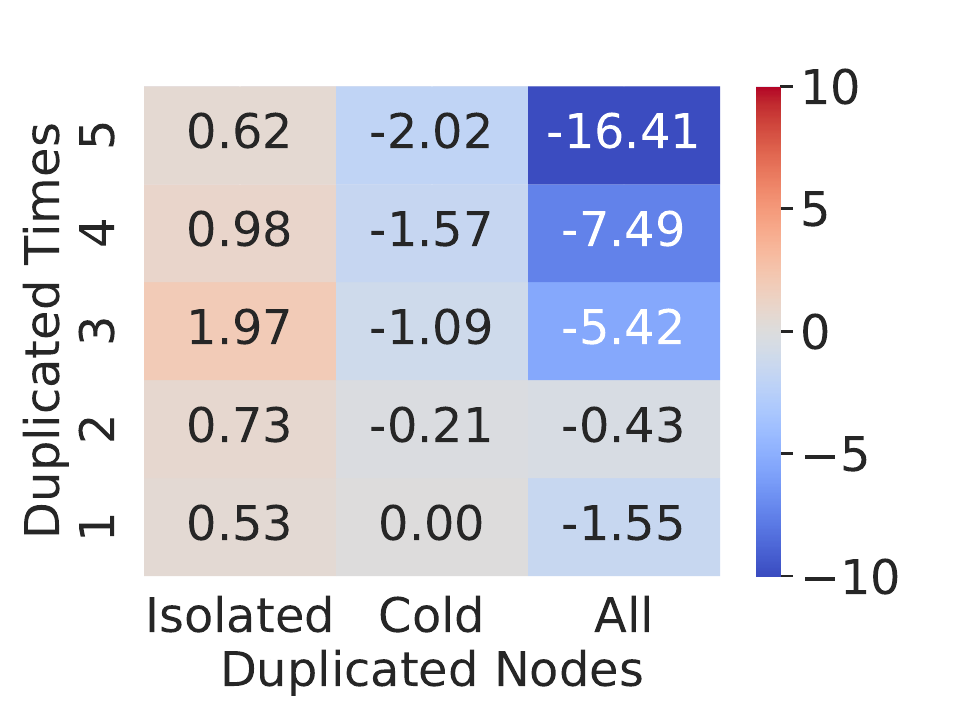}\label{fig:cold}}
    \subfigure[\overall]
    {\includegraphics[width=0.32\linewidth]{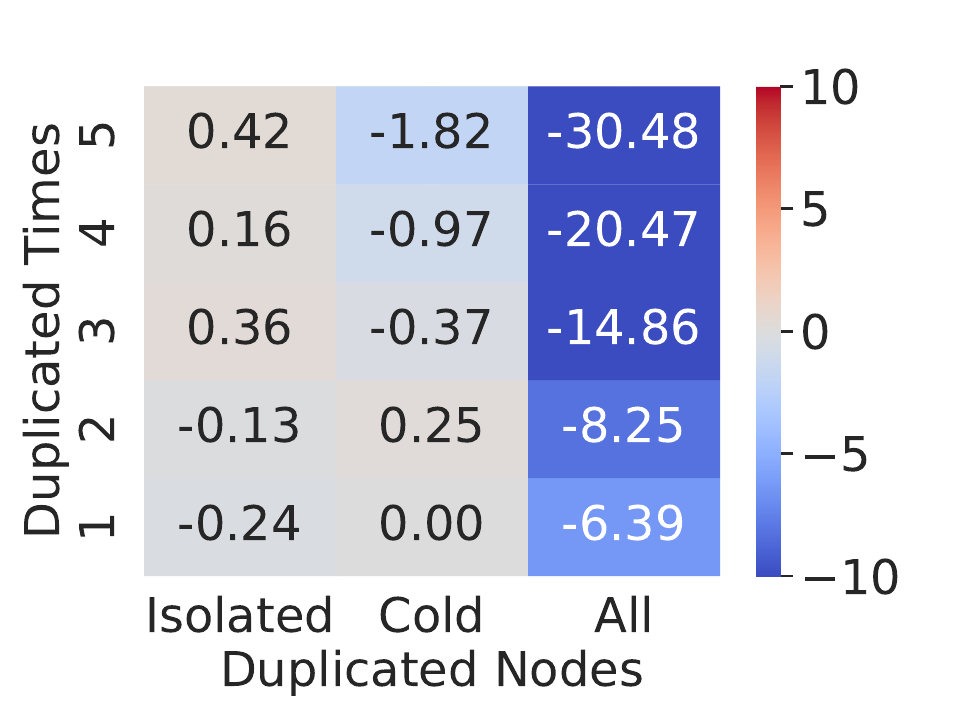}\label{fig:overall}}
    \vspace{-0.2in}
    \caption{Ablation study on duplication frequency and node types of \ours. (a), (b), (c) show the performance for \isolated nodes, \cold nodes, and \overall settings, respectively. The values in each block represent the performance differences compared to the baseline setting of duplicating cold nodes once.}
    \label{fig:ablation_duplicate}
\end{figure}

In our experiments, we duplicate the cold nodes once and add one edge for each cold node in \ours. \cref{fig:ablation_duplicate} presents the results of our ablation study on \citeseer dataset, which examines how varying the duplication frequency and the types of duplicated nodes affect the performance of \ours across the \isolated, \cold, and \overall settings. The numbers in each block represent the performance differences relative to the baseline of duplicating cold nodes once. From the results, we observe that increasing the duplication frequency does not consistently lead to performance improvements across all settings. Notably, duplicating all nodes multiple times significantly enhances the performance of \isolated nodes. However, this approach also introduces a large number of isolated nodes into the graph, which negatively impacts the overall performance. Consequently, duplicating only cold nodes once emerges as the most effective strategy, as it consistently maintains strong performance across all settings. Further analysis of the impact of different duplicated node types is provided in \cref{sec:nodetypes}.

\subsection{Analyzing Performance Gains of \ours on \warm Nodes}
\label{appendix:warm-warm analysis}
\begin{table}[]
\centering
\small
\caption{Distribution and AUC performance of testing \warm-\warm and \warm-Cold links. \ours improves Warm-Cold link performance while maintaining Warm-Warm link performance.}
\label{tab:warmwarm}
\scalebox{0.9}{
\begin{tabular}{l|cccc|cccc}
\toprule
\multicolumn{1}{c|}{\multirow{2}{*}{}} & \multicolumn{4}{c|}{\warm-\warm}                            & \multicolumn{4}{c}{\warm-Cold}                            \\ 
\multicolumn{1}{c|}{}                  & Number   & GSage                     & \ourslight & \ours & Number  & GSage                     & \ourslight & \ours \\\midrule
\cora                                  & 157738   & \multicolumn{1}{l}{\ms{94.92}{0.31}} & \ms{95.17}{0.19}      & \textbf{\ms{95.18}{0.18}}   & 16759   & \multicolumn{1}{l}{\ms{77.06}{1.40}} & \textbf{\ms{81.41}{1.18}}      & \ms{80.51}{1.72}   \\
\citeseer                              & 63266    & \multicolumn{1}{l}{\textbf{\ms{97.21}{0.09}}} & \ms{97.06}{0.21}     & \ms{97.02}{0.12}   & 24020   & \ms{85.40}{0.78}                     & \ms{87.96}{0.79}      & \textbf{\ms{88.40}{0.92}}    \\
\cs                                    & 4209161  & \ms{98.31}{0.03}                     & \ms{98.30}{0.02}      & \textbf{\ms{98.42}{0.02}}   & 91458   & \ms{87.92}{0.19}                     & \textbf{\ms{91.47}{0.35}}      & \ms{90.44}{0.84}   \\
\physics                               & 11462743 & \multicolumn{1}{l}{\ms{99.01}{0.01}} & \ms{99.01}{0.02}      & \textbf{\ms{99.02}{0.00}}   & 103174  & \ms{86.21}{0.33}                     & \ms{89.94}{0.31}      & \textbf{\ms{90.23}{0.51}}   \\
\photos                                & 2984253  & \ms{97.85}{0.06}                     & \textbf{\ms{98.03}{0.04}}      & \ms{97.87}{0.02}   & 104737  & \ms{59.80}{1.33}                     & \textbf{\ms{68.11}{0.43}}      & \ms{64.32}{0.73}   \\
\computers                             & 5417165  & \ms{97.58}{0.07}                     & \textbf{\ms{97.60}{0.08}}       & \ms{97.54}{0.09}   & 217090  & \ms{46.49}{0.75}                     & \ms{57.32}{0.99}      & \textbf{\ms{57.63}{0.49}}   \\
\igbtiny                              & 6899924  & \ms{98.70}{0.00}                     & \textbf{\ms{98.71}{0.02}}      & \ms{98.64}{0.01}   & 1372994 & \ms{97.14}{0.10}                     & \textbf{\ms{98.63}{0.42}}      & \ms{98.23}{0.06}   \\
 \bottomrule
\end{tabular}}
\end{table}

To better understand the performance improvements of \ours on \warm nodes, we first analyzed the distribution of \warm-\warm and \warm-Cold links in the testing set. Our analysis reveals that, across all datasets, the number of \warm-\warm links consistently exceeds that of \warm-Cold links, as shown in~\cref{tab:warmwarm}. This observation indicates that \warm-Cold links do not overwhelmingly influence the overall performance of \warm nodes. To further investigate, we conducted experiments comparing the performance of our method on these two types of links. The results demonstrate that our approach consistently enhances performance on \warm-Cold links while maintaining strong performance on \warm-\warm links. These findings confirm that our method does not negatively impact the learning ability of \warm nodes. Instead, the observed performance gains primarily stem from improved learning on Cold nodes, as previously discussed in~\cref{sec:base}.


\section{Conclusion}
\label{sec:conclusion}
GNNs in LP encounter difficulties when dealing with cold nodes that lack sufficient or absent neighbors. To address this challenge, we presented a simple yet effective augmentation method (\ours) specifically tailored for the cold-start LP problem, which can effectively enhance the prediction capabilities of GNNs for cold nodes while maintaining overall performance. Extensive evaluations demonstrated that both \ours and its lightweight variant, \ourslight, consistently outperformed baselines on both cold node and warm node settings across 7 benchmark datasets. \ours also achieved better runtime efficiency compared to the augmentation baselines.

\clearpage

\section*{Acknowledgements}
We are grateful for all the insightful comments and constructive suggestions provided by the reviewers. This research was partially conducted during Zhichun’s internship at Snap Inc. and was supported by the National Science Foundation (NSF) through the Center for Computer Assisted Synthesis (C-CAS), under grant number CHE-2202693.

\bibliography{ref}

\begin{thebibliography}{86}
\providecommand{\natexlab}[1]{#1}
\providecommand{\url}[1]{\texttt{#1}}
\expandafter\ifx\csname urlstyle\endcsname\relax
  \providecommand{\doi}[1]{doi: #1}\else
  \providecommand{\doi}{doi: \begingroup \urlstyle{rm}\Url}\fi

\bibitem[Allen-Zhu \& Li(2020)Allen-Zhu and Li]{allen2020towards}
Zeyuan Allen-Zhu and Yuanzhi Li.
\newblock Towards understanding ensemble, knowledge distillation and self-distillation in deep learning.
\newblock \emph{arXiv preprint arXiv:2012.09816}, 2020.

\bibitem[Berg et~al.(2017)Berg, Kipf, and Welling]{berg2017graph}
Rianne van~den Berg, Thomas~N Kipf, and Max Welling.
\newblock Graph convolutional matrix completion.
\newblock \emph{arXiv preprint arXiv:1706.02263}, 2017.

\bibitem[Cai \& Ji(2020)Cai and Ji]{cai2020multi}
Lei Cai and Shuiwang Ji.
\newblock A multi-scale approach for graph link prediction.
\newblock In \emph{Proceedings of the AAAI conference on artificial intelligence}, 2020.

\bibitem[Cai et~al.(2021)Cai, Li, Wang, and Ji]{cai2021line}
Lei Cai, Jundong Li, Jie Wang, and Shuiwang Ji.
\newblock Line graph neural networks for link prediction.
\newblock \emph{IEEE Transactions on Pattern Analysis and Machine Intelligence}, 2021.

\bibitem[Cai et~al.(2019)Cai, Yan, Mai, Janowicz, and Zhu]{cai2019transgcn}
Ling Cai, Bo~Yan, Gengchen Mai, Krzysztof Janowicz, and Rui Zhu.
\newblock Transgcn: Coupling transformation assumptions with graph convolutional networks for link prediction.
\newblock In \emph{Proceedings of the 10th international conference on knowledge capture}, pp.\  131--138, 2019.

\bibitem[Chamberlain et~al.(2022)Chamberlain, Shirobokov, Rossi, Frasca, Markovich, Hammerla, Bronstein, and Hansmire]{chamberlain2022graph}
Benjamin~Paul Chamberlain, Sergey Shirobokov, Emanuele Rossi, Fabrizio Frasca, Thomas Markovich, Nils Hammerla, Michael~M Bronstein, and Max Hansmire.
\newblock Graph neural networks for link prediction with subgraph sketching.
\newblock \emph{arXiv preprint arXiv:2209.15486}, 2022.

\bibitem[Chen et~al.(2020)Chen, Xiao, Li, Ye, Sun, and Deng]{chen2020esam}
Zhihong Chen, Rong Xiao, Chenliang Li, Gangfeng Ye, Haochuan Sun, and Hongbo Deng.
\newblock Esam: Discriminative domain adaptation with non-displayed items to improve long-tail performance.
\newblock In \emph{Proceedings of the 43rd International ACM SIGIR Conference on Research and Development in Information Retrieval}, pp.\  579--588, 2020.

\bibitem[Chien et~al.(2021)Chien, Chang, Hsieh, Yu, Zhang, Milenkovic, and Dhillon]{chien2021node}
Eli Chien, Wei-Cheng Chang, Cho-Jui Hsieh, Hsiang-Fu Yu, Jiong Zhang, Olgica Milenkovic, and Inderjit~S Dhillon.
\newblock Node feature extraction by self-supervised multi-scale neighborhood prediction.
\newblock \emph{arXiv preprint arXiv:2111.00064}, 2021.

\bibitem[Clauset et~al.(2009)Clauset, Shalizi, and Newman]{clauset2009power}
Aaron Clauset, Cosma~Rohilla Shalizi, and Mark~EJ Newman.
\newblock Power-law distributions in empirical data.
\newblock \emph{SIAM review}, 2009.

\bibitem[Davidson et~al.(2018)Davidson, Falorsi, De~Cao, Kipf, and Tomczak]{davidson2018hyperspherical}
Tim~R Davidson, Luca Falorsi, Nicola De~Cao, Thomas Kipf, and Jakub~M Tomczak.
\newblock Hyperspherical variational auto-encoders.
\newblock \emph{arXiv preprint arXiv:1804.00891}, 2018.

\bibitem[Ding et~al.(2021)Ding, Ma, Deoras, Wang, and Wang]{ding2021zero}
Hao Ding, Yifei Ma, Anoop Deoras, Yuyang Wang, and Hao Wang.
\newblock Zero-shot recommender systems.
\newblock \emph{arXiv preprint arXiv:2105.08318}, 2021.

\bibitem[Ding et~al.(2022)Ding, Xu, Tong, and Liu]{ding2022data}
Kaize Ding, Zhe Xu, Hanghang Tong, and Huan Liu.
\newblock Data augmentation for deep graph learning: A survey.
\newblock \emph{ACM SIGKDD Explorations Newsletter}, 2022.

\bibitem[Dong et~al.(2022)Dong, Tian, Guo, Yang, and Chawla]{dong2022fakeedge}
Kaiwen Dong, Yijun Tian, Zhichun Guo, Yang Yang, and Nitesh Chawla.
\newblock Fakeedge: Alleviate dataset shift in link prediction.
\newblock In \emph{Learning on Graphs Conference}, pp.\  56--1. PMLR, 2022.

\bibitem[Dwivedi \& Bresson(2020)Dwivedi and Bresson]{dwivedi2020generalization}
Vijay~Prakash Dwivedi and Xavier Bresson.
\newblock A generalization of transformer networks to graphs.
\newblock \emph{arXiv preprint arXiv:2012.09699}, 2020.

\bibitem[Fan et~al.(2022)Fan, Liu, Jin, Zhao, Tang, and Li]{fan2022graph}
Wenqi Fan, Xiaorui Liu, Wei Jin, Xiangyu Zhao, Jiliang Tang, and Qing Li.
\newblock Graph trend filtering networks for recommendation.
\newblock In \emph{Proceedings of the 45th International ACM SIGIR Conference on Research and Development in Information Retrieval}, 2022.

\bibitem[Feng et~al.(2020)Feng, Zhang, Dong, Han, Luan, Xu, Yang, Kharlamov, and Tang]{feng2020graph}
Wenzheng Feng, Jie Zhang, Yuxiao Dong, Yu~Han, Huanbo Luan, Qian Xu, Qiang Yang, Evgeny Kharlamov, and Jie Tang.
\newblock Graph random neural networks for semi-supervised learning on graphs.
\newblock \emph{Advances in neural information processing systems}, 33:\penalty0 22092--22103, 2020.

\bibitem[Fey \& Lenssen(2019)Fey and Lenssen]{fey2019fast}
Matthias Fey and Jan~Eric Lenssen.
\newblock Fast graph representation learning with pytorch geometric.
\newblock \emph{arXiv preprint arXiv:1903.02428}, 2019.

\bibitem[Gilmer et~al.(2017)Gilmer, Schoenholz, Riley, Vinyals, and Dahl]{gilmer2017neural}
Justin Gilmer, Samuel~S Schoenholz, Patrick~F Riley, Oriol Vinyals, and George~E Dahl.
\newblock Neural message passing for quantum chemistry.
\newblock In \emph{International conference on machine learning}. PMLR, 2017.

\bibitem[Guo et~al.(2022)Guo, Shiao, Zhang, Liu, Chawla, Shah, and Zhao]{guo2022linkless}
Zhichun Guo, William Shiao, Shichang Zhang, Yozen Liu, Nitesh Chawla, Neil Shah, and Tong Zhao.
\newblock Linkless link prediction via relational distillation.
\newblock \emph{arXiv preprint arXiv:2210.05801}, 2022.

\bibitem[Hamilton et~al.(2017)Hamilton, Ying, and Leskovec]{hamilton2017inductive}
Will Hamilton, Zhitao Ying, and Jure Leskovec.
\newblock Inductive representation learning on large graphs.
\newblock \emph{Advances in neural information processing systems}, 2017.

\bibitem[Han et~al.(2022)Han, Zhao, Liu, Hu, and Shah]{han2022mlpinit}
Xiaotian Han, Tong Zhao, Yozen Liu, Xia Hu, and Neil Shah.
\newblock Mlpinit: Embarrassingly simple gnn training acceleration with mlp initialization.
\newblock \emph{arXiv preprint arXiv:2210.00102}, 2022.

\bibitem[Hao et~al.(2021)Hao, Zhang, Yin, Li, and Chen]{hao2021pre}
Bowen Hao, Jing Zhang, Hongzhi Yin, Cuiping Li, and Hong Chen.
\newblock Pre-training graph neural networks for cold-start users and items representation.
\newblock In \emph{Proceedings of the 14th ACM International Conference on Web Search and Data Mining}, 2021.

\bibitem[He et~al.(2020)He, Deng, Wang, Li, Zhang, and Wang]{he2020lightgcn}
Xiangnan He, Kuan Deng, Xiang Wang, Yan Li, Yongdong Zhang, and Meng Wang.
\newblock Lightgcn: Simplifying and powering graph convolution network for recommendation.
\newblock In \emph{Proceedings of the 43rd International ACM SIGIR conference on research and development in Information Retrieval}, 2020.

\bibitem[Hu et~al.(2020)Hu, Fey, Zitnik, Dong, Ren, Liu, Catasta, and Leskovec]{hu2020open}
Weihua Hu, Matthias Fey, Marinka Zitnik, Yuxiao Dong, Hongyu Ren, Bowen Liu, Michele Catasta, and Jure Leskovec.
\newblock Open graph benchmark: Datasets for machine learning on graphs.
\newblock \emph{Advances in neural information processing systems}, 2020.

\bibitem[Hu et~al.(2022)Hu, Cao, Huang, Huang, Subbian, and Leskovec]{hu2022tuneup}
Weihua Hu, Kaidi Cao, Kexin Huang, Edward~W Huang, Karthik Subbian, and Jure Leskovec.
\newblock Tuneup: A training strategy for improving generalization of graph neural networks.
\newblock \emph{arXiv preprint arXiv:2210.14843}, 2022.

\bibitem[Huang et~al.(2021)Huang, Xu, Xu, Dai, Xia, Lu, Bo, Xing, Lai, and Ye]{huang2021knowledge}
Chao Huang, Huance Xu, Yong Xu, Peng Dai, Lianghao Xia, Mengyin Lu, Liefeng Bo, Hao Xing, Xiaoping Lai, and Yanfang Ye.
\newblock Knowledge-aware coupled graph neural network for social recommendation.
\newblock In \emph{Proceedings of the AAAI conference on artificial intelligence}, volume~35, pp.\  4115--4122, 2021.

\bibitem[Kang et~al.(2019)Kang, Xie, Rohrbach, Yan, Gordo, Feng, and Kalantidis]{kang2019decoupling}
Bingyi Kang, Saining Xie, Marcus Rohrbach, Zhicheng Yan, Albert Gordo, Jiashi Feng, and Yannis Kalantidis.
\newblock Decoupling representation and classifier for long-tailed recognition.
\newblock \emph{arXiv preprint arXiv:1910.09217}, 2019.

\bibitem[Khatua et~al.(2023)Khatua, Mailthody, Taleka, Ma, Song, and Hwu]{igbdatasets}
Arpandeep Khatua, Vikram~Sharma Mailthody, Bhagyashree Taleka, Tengfei Ma, Xiang Song, and Wen-mei Hwu.
\newblock Igb: Addressing the gaps in labeling, features, heterogeneity, and size of public graph datasets for deep learning research, 2023.
\newblock URL \url{https://arxiv.org/abs/2302.13522}.

\bibitem[Kipf \& Welling(2016{\natexlab{a}})Kipf and Welling]{kipf2016semi}
Thomas~N Kipf and Max Welling.
\newblock Semi-supervised classification with graph convolutional networks.
\newblock \emph{arXiv preprint arXiv:1609.02907}, 2016{\natexlab{a}}.

\bibitem[Kipf \& Welling(2016{\natexlab{b}})Kipf and Welling]{kipf2016variational}
Thomas~N Kipf and Max Welling.
\newblock Variational graph auto-encoders.
\newblock \emph{arXiv preprint arXiv:1611.07308}, 2016{\natexlab{b}}.

\bibitem[Kirkpatrick et~al.(2017)Kirkpatrick, Pascanu, Rabinowitz, Veness, Desjardins, Rusu, Milan, Quan, Ramalho, Grabska-Barwinska, et~al.]{kirkpatrick2017overcoming}
James Kirkpatrick, Razvan Pascanu, Neil Rabinowitz, Joel Veness, Guillaume Desjardins, Andrei~A Rusu, Kieran Milan, John Quan, Tiago Ramalho, Agnieszka Grabska-Barwinska, et~al.
\newblock Overcoming catastrophic forgetting in neural networks.
\newblock \emph{Proceedings of the national academy of sciences}, 114\penalty0 (13):\penalty0 3521--3526, 2017.

\bibitem[Kov{\'a}cs et~al.(2019)Kov{\'a}cs, Luck, Spirohn, Wang, Pollis, Schlabach, Bian, Kim, Kishore, Hao, et~al.]{kovacs2019network}
Istv{\'a}n~A Kov{\'a}cs, Katja Luck, Kerstin Spirohn, Yang Wang, Carl Pollis, Sadie Schlabach, Wenting Bian, Dae-Kyum Kim, Nishka Kishore, Tong Hao, et~al.
\newblock Network-based prediction of protein interactions.
\newblock \emph{Nature communications}, 2019.

\bibitem[Li et~al.(2023)Li, Shomer, Ding, Wang, Ma, Shah, Tang, and Yin]{li2023message}
Juanhui Li, Harry Shomer, Jiayuan Ding, Yiqi Wang, Yao Ma, Neil Shah, Jiliang Tang, and Dawei Yin.
\newblock Are message passing neural networks really helpful for knowledge graph completion?
\newblock \emph{ACL}, 2023.

\bibitem[Liao et~al.(2023)Liao, Li, Chen, Wu, Bian, and Zheng]{liao2023sailor}
Jie Liao, Jintang Li, Liang Chen, Bingzhe Wu, Yatao Bian, and Zibin Zheng.
\newblock Sailor: Structural augmentation based tail node representation learning.
\newblock In \emph{Proceedings of the 32nd ACM International Conference on Information and Knowledge Management}, pp.\  1389--1399, 2023.

\bibitem[Liben-Nowell \& Kleinberg(2007)Liben-Nowell and Kleinberg]{liben2007link}
David Liben-Nowell and Jon Kleinberg.
\newblock The link-prediction problem for social networks.
\newblock \emph{Journal of the American Society for Information Science and Technology}, 2007.

\bibitem[Lin et~al.(2017)Lin, Goyal, Girshick, He, and Doll{\'a}r]{lin2017focal}
Tsung-Yi Lin, Priya Goyal, Ross Girshick, Kaiming He, and Piotr Doll{\'a}r.
\newblock Focal loss for dense object detection.
\newblock In \emph{Proceedings of the IEEE international conference on computer vision}, pp.\  2980--2988, 2017.

\bibitem[Liu et~al.(2022{\natexlab{a}})Liu, Zhao, Xu, Luo, and Jiang]{liu2022graph}
Gang Liu, Tong Zhao, Jiaxin Xu, Tengfei Luo, and Meng Jiang.
\newblock Graph rationalization with environment-based augmentations.
\newblock In \emph{Proceedings of the 28th ACM SIGKDD Conference on Knowledge Discovery and Data Mining}, pp.\  1069--1078, 2022{\natexlab{a}}.

\bibitem[Liu \& Zheng(2020)Liu and Zheng]{liu2020long}
Siyi Liu and Yujia Zheng.
\newblock Long-tail session-based recommendation.
\newblock In \emph{Proceedings of the 14th ACM Conference on Recommender Systems}, pp.\  509--514, 2020.

\bibitem[Liu et~al.(2022{\natexlab{b}})Liu, Ying, Dong, Li, Xu, Rong, Zhao, Huang, and Wu]{liu2022local}
Songtao Liu, Rex Ying, Hanze Dong, Lanqing Li, Tingyang Xu, Yu~Rong, Peilin Zhao, Junzhou Huang, and Dinghao Wu.
\newblock Local augmentation for graph neural networks.
\newblock In \emph{International Conference on Machine Learning}, pp.\  14054--14072. PMLR, 2022{\natexlab{b}}.

\bibitem[Liu et~al.(2020)Liu, Zhang, Fang, Zhang, and Hoi]{liu2020towards}
Zemin Liu, Wentao Zhang, Yuan Fang, Xinming Zhang, and Steven~CH Hoi.
\newblock Towards locality-aware meta-learning of tail node embeddings on networks.
\newblock In \emph{Proceedings of the 29th ACM International Conference on Information \& Knowledge Management}, pp.\  975--984, 2020.

\bibitem[Liu et~al.(2021)Liu, Nguyen, and Fang]{liu2021tail}
Zemin Liu, Trung-Kien Nguyen, and Yuan Fang.
\newblock Tail-gnn: Tail-node graph neural networks.
\newblock In \emph{Proceedings of the 27th ACM SIGKDD Conference on Knowledge Discovery \& Data Mining}, 2021.

\bibitem[Liu et~al.(2023)Liu, Nguyen, and Fang]{liu2023generalized}
Zemin Liu, Trung-Kien Nguyen, and Yuan Fang.
\newblock On generalized degree fairness in graph neural networks.
\newblock \emph{arXiv preprint arXiv:2302.03881}, 2023.

\bibitem[Lu et~al.(2020)Lu, Fang, and Shi]{lu2020meta}
Yuanfu Lu, Yuan Fang, and Chuan Shi.
\newblock Meta-learning on heterogeneous information networks for cold-start recommendation.
\newblock In \emph{Proceedings of the 26th ACM SIGKDD International Conference on Knowledge Discovery \& Data Mining}, 2020.

\bibitem[Luo et~al.(2022)Luo, McThrow, Au, Komikado, Uchino, Maruhash, and Ji]{luo2022automated}
Youzhi Luo, Michael McThrow, Wing~Yee Au, Tao Komikado, Kanji Uchino, Koji Maruhash, and Shuiwang Ji.
\newblock Automated data augmentations for graph classification.
\newblock \emph{arXiv preprint arXiv:2202.13248}, 2022.

\bibitem[Park et~al.(2021)Park, Lee, Kim, Park, Jeong, Kim, Ha, and Kim]{park2021metropolis}
Hyeonjin Park, Seunghun Lee, Sihyeon Kim, Jinyoung Park, Jisu Jeong, Kyung-Min Kim, Jung-Woo Ha, and Hyunwoo~J Kim.
\newblock Metropolis-hastings data augmentation for graph neural networks.
\newblock \emph{Advances in Neural Information Processing Systems}, 34:\penalty0 19010--19020, 2021.

\bibitem[Pei et~al.(2020)Pei, Wei, Chang, Lei, and Yang]{pei2020geom}
Hongbin Pei, Bingzhe Wei, Kevin Chen-Chuan Chang, Yu~Lei, and Bo~Yang.
\newblock Geom-gcn: Geometric graph convolutional networks.
\newblock \emph{arXiv preprint arXiv:2002.05287}, 2020.

\bibitem[Provost(2000)]{provost2000machine}
Foster Provost.
\newblock Machine learning from imbalanced data sets 101.
\newblock 2000.

\bibitem[Ren et~al.(2020)Ren, Yu, Ma, Zhao, Yi, et~al.]{ren2020balanced}
Jiawei Ren, Cunjun Yu, Xiao Ma, Haiyu Zhao, Shuai Yi, et~al.
\newblock Balanced meta-softmax for long-tailed visual recognition.
\newblock \emph{Advances in neural information processing systems}, 33:\penalty0 4175--4186, 2020.

\bibitem[Rong et~al.(2019)Rong, Huang, Xu, and Huang]{rong2019dropedge}
Yu~Rong, Wenbing Huang, Tingyang Xu, and Junzhou Huang.
\newblock Dropedge: Towards deep graph convolutional networks on node classification.
\newblock \emph{arXiv preprint arXiv:1907.10903}, 2019.

\bibitem[Sankar et~al.(2021)Sankar, Liu, Yu, and Shah]{sankar2021graph}
Aravind Sankar, Yozen Liu, Jun Yu, and Neil Shah.
\newblock Graph neural networks for friend ranking in large-scale social platforms.
\newblock In \emph{Proceedings of the Web Conference 2021}, 2021.

\bibitem[Schlichtkrull et~al.(2018)Schlichtkrull, Kipf, Bloem, Berg, Titov, and Welling]{schlichtkrull2018modeling}
Michael Schlichtkrull, Thomas~N Kipf, Peter Bloem, Rianne van~den Berg, Ivan Titov, and Max Welling.
\newblock Modeling relational data with graph convolutional networks.
\newblock In \emph{European semantic web conference}, pp.\  593--607. Springer, 2018.

\bibitem[Shiao et~al.(2022)Shiao, Guo, Zhao, Papalexakis, Liu, and Shah]{shiao2022link}
William Shiao, Zhichun Guo, Tong Zhao, Evangelos~E Papalexakis, Yozen Liu, and Neil Shah.
\newblock Link prediction with non-contrastive learning.
\newblock \emph{arXiv preprint arXiv:2211.14394}, 2022.

\bibitem[Stanfield et~al.(2017)Stanfield, Co{\c{s}}kun, and Koyut{\"u}rk]{stanfield2017drug}
Zachary Stanfield, Mustafa Co{\c{s}}kun, and Mehmet Koyut{\"u}rk.
\newblock Drug response prediction as a link prediction problem.
\newblock \emph{Scientific reports}, 2017.

\bibitem[Tan et~al.(2020)Tan, Wang, Li, Li, Ouyang, Yin, and Yan]{tan2020equalization}
Jingru Tan, Changbao Wang, Buyu Li, Quanquan Li, Wanli Ouyang, Changqing Yin, and Junjie Yan.
\newblock Equalization loss for long-tailed object recognition.
\newblock In \emph{Proceedings of the IEEE/CVF conference on computer vision and pattern recognition}, pp.\  11662--11671, 2020.

\bibitem[Tang et~al.(2020{\natexlab{a}})Tang, Huang, and Zhang]{tang2020long}
Kaihua Tang, Jianqiang Huang, and Hanwang Zhang.
\newblock Long-tailed classification by keeping the good and removing the bad momentum causal effect.
\newblock \emph{Advances in Neural Information Processing Systems}, 33:\penalty0 1513--1524, 2020{\natexlab{a}}.

\bibitem[Tang et~al.(2020{\natexlab{b}})Tang, Yao, Sun, Wang, Tang, Aggarwal, Mitra, and Wang]{tang2020investigating}
Xianfeng Tang, Huaxiu Yao, Yiwei Sun, Yiqi Wang, Jiliang Tang, Charu Aggarwal, Prasenjit Mitra, and Suhang Wang.
\newblock Investigating and mitigating degree-related biases in graph convoltuional networks.
\newblock In \emph{Proceedings of the 29th ACM International Conference on Information \& Knowledge Management}, 2020{\natexlab{b}}.

\bibitem[Tang et~al.(2022)Tang, Liu, He, Wang, and Shah]{tang2022friend}
Xianfeng Tang, Yozen Liu, Xinran He, Suhang Wang, and Neil Shah.
\newblock Friend story ranking with edge-contextual local graph convolutions.
\newblock In \emph{Proceedings of the Fifteenth ACM International Conference on Web Search and Data Mining}, 2022.

\bibitem[Trouillon et~al.(2016)Trouillon, Welbl, Riedel, Gaussier, and Bouchard]{trouillon2016complex}
Th{\'e}o Trouillon, Johannes Welbl, Sebastian Riedel, {\'E}ric Gaussier, and Guillaume Bouchard.
\newblock Complex embeddings for simple link prediction.
\newblock In \emph{International conference on machine learning}. PMLR, 2016.

\bibitem[Valkanas et~al.(2024)Valkanas, Wang, Zhang, and Coates]{valkanas2024personalized}
Antonios Valkanas, Yuening Wang, Yingxue Zhang, and Mark Coates.
\newblock Personalized negative reservoir for incremental learning in recommender systems.
\newblock \emph{arXiv preprint arXiv:2403.03993}, 2024.

\bibitem[Vashishth et~al.(2020)Vashishth, Sanyal, Nitin, and Talukdar]{vashishth2020composition}
Shikhar Vashishth, Soumya Sanyal, Vikram Nitin, and Partha Talukdar.
\newblock Composition-based multi-relational graph convolutional networks.
\newblock In \emph{International Conference on Learning Representations}, 2020.

\bibitem[Veli{\v{c}}kovi{\'c} et~al.(2017)Veli{\v{c}}kovi{\'c}, Cucurull, Casanova, Romero, Lio, and Bengio]{velivckovic2017graph}
Petar Veli{\v{c}}kovi{\'c}, Guillem Cucurull, Arantxa Casanova, Adriana Romero, Pietro Lio, and Yoshua Bengio.
\newblock Graph attention networks.
\newblock \emph{arXiv preprint arXiv:1710.10903}, 2017.

\bibitem[Volkovs et~al.(2017)Volkovs, Yu, and Poutanen]{volkovs2017dropoutnet}
Maksims Volkovs, Guangwei Yu, and Tomi Poutanen.
\newblock Dropoutnet: Addressing cold start in recommender systems.
\newblock \emph{Advances in neural information processing systems}, 30, 2017.

\bibitem[Wang et~al.(2022)Wang, Wang, Shi, and Song]{wang2022uncovering}
Ruijia Wang, Xiao Wang, Chuan Shi, and Le~Song.
\newblock Uncovering the structural fairness in graph contrastive learning.
\newblock \emph{Advances in neural information processing systems}, 35:\penalty0 32465--32473, 2022.

\bibitem[Wang et~al.(2020)Wang, Lei, and Li]{wang2020neighborhood}
Zhitao Wang, Yu~Lei, and Wenjie Li.
\newblock Neighborhood attention networks with adversarial learning for link prediction.
\newblock \emph{IEEE Transactions on Neural Networks and Learning Systems}, 32\penalty0 (8):\penalty0 3653--3663, 2020.

\bibitem[Wang et~al.(2021)Wang, Zhou, Hong, Zou, Su, and Chen]{wang2021pairwise}
Zhitao Wang, Yong Zhou, Litao Hong, Yuanhang Zou, Hanjing Su, and Shouzhi Chen.
\newblock Pairwise learning for neural link prediction.
\newblock \emph{arXiv preprint arXiv:2112.02936}, 2021.

\bibitem[Wu et~al.(2019)Wu, He, and Xu]{wu2019net}
Jun Wu, Jingrui He, and Jiejun Xu.
\newblock Net: Degree-specific graph neural networks for node and graph classification.
\newblock In \emph{Proceedings of the 25th ACM SIGKDD International Conference on Knowledge Discovery \& Data Mining}, pp.\  406--415, 2019.

\bibitem[Wu et~al.(2022)Wu, Zhang, Yan, and Wipf]{wu2022handling}
Qitian Wu, Hengrui Zhang, Junchi Yan, and David Wipf.
\newblock Handling distribution shifts on graphs: An invariance perspective.
\newblock \emph{arXiv preprint arXiv:2202.02466}, 2022.

\bibitem[Wu et~al.(2020)Wu, Pan, Chen, Long, Zhang, and Philip]{wu2020comprehensive}
Zonghan Wu, Shirui Pan, Fengwen Chen, Guodong Long, Chengqi Zhang, and S~Yu Philip.
\newblock A comprehensive survey on graph neural networks.
\newblock \emph{IEEE transactions on neural networks and learning systems}, 2020.

\bibitem[Xu et~al.(2018)Xu, Li, Tian, Sonobe, Kawarabayashi, and Jegelka]{xu2018representation}
Keyulu Xu, Chengtao Li, Yonglong Tian, Tomohiro Sonobe, Ken-ichi Kawarabayashi, and Stefanie Jegelka.
\newblock Representation learning on graphs with jumping knowledge networks.
\newblock In \emph{International conference on machine learning}, pp.\  5453--5462. PMLR, 2018.

\bibitem[Xu et~al.(2020)Xu, Zhang, Guo, Guo, Tang, and Coates]{xu2020graphsail}
Yishi Xu, Yingxue Zhang, Wei Guo, Huifeng Guo, Ruiming Tang, and Mark Coates.
\newblock Graphsail: Graph structure aware incremental learning for recommender systems.
\newblock In \emph{Proceedings of the 29th ACM International Conference on Information \& Knowledge Management}, pp.\  2861--2868, 2020.

\bibitem[Yan et~al.(2021)Yan, Ma, Gao, Tang, and Chen]{yan2021link}
Zuoyu Yan, Tengfei Ma, Liangcai Gao, Zhi Tang, and Chao Chen.
\newblock Link prediction with persistent homology: An interactive view.
\newblock In \emph{International conference on machine learning}, pp.\  11659--11669. PMLR, 2021.

\bibitem[Yang et~al.(2021)Yang, Xiao, Ma, Glass, and Sun]{yang2021safedrug}
Chaoqi Yang, Cao Xiao, Fenglong Ma, Lucas Glass, and Jimeng Sun.
\newblock Safedrug: Dual molecular graph encoders for recommending effective and safe drug combinations.
\newblock In \emph{IJCAI}, pp.\  3735--3741, 2021.

\bibitem[Yang et~al.(2016)Yang, Cohen, and Salakhudinov]{yang2016revisiting}
Zhilin Yang, William Cohen, and Ruslan Salakhudinov.
\newblock Revisiting semi-supervised learning with graph embeddings.
\newblock In \emph{International conference on machine learning}, pp.\  40--48. PMLR, 2016.

\bibitem[Ying et~al.(2018)Ying, He, Chen, Eksombatchai, Hamilton, and Leskovec]{ying2018graph}
Rex Ying, Ruining He, Kaifeng Chen, Pong Eksombatchai, William~L Hamilton, and Jure Leskovec.
\newblock Graph convolutional neural networks for web-scale recommender systems.
\newblock In \emph{Proceedings of the 24th ACM SIGKDD international conference on knowledge discovery \& data mining}, 2018.

\bibitem[Yun et~al.(2021)Yun, Kim, Lee, Kang, and Kim]{yun2021neo}
Seongjun Yun, Seoyoon Kim, Junhyun Lee, Jaewoo Kang, and Hyunwoo~J Kim.
\newblock Neo-gnns: Neighborhood overlap-aware graph neural networks for link prediction.
\newblock \emph{Advances in Neural Information Processing Systems}, 2021.

\bibitem[Zhang et~al.(2020)Zhang, Yao, Huang, Jiang, Li, and Chawla]{zhang2020few}
Chuxu Zhang, Huaxiu Yao, Chao Huang, Meng Jiang, Zhenhui Li, and Nitesh~V Chawla.
\newblock Few-shot knowledge graph completion.
\newblock In \emph{Proceedings of the AAAI Conference on Artificial Intelligence}, 2020.

\bibitem[Zhang \& Chen(2018)Zhang and Chen]{zhang2018link}
Muhan Zhang and Yixin Chen.
\newblock Link prediction based on graph neural networks.
\newblock \emph{Advances in neural information processing systems}, 2018.

\bibitem[Zhang et~al.(2021)Zhang, Li, Xia, Wang, and Jin]{zhang2021labeling}
Muhan Zhang, Pan Li, Yinglong Xia, Kai Wang, and Long Jin.
\newblock Labeling trick: A theory of using graph neural networks for multi-node representation learning.
\newblock \emph{Advances in Neural Information Processing Systems}, 2021.

\bibitem[Zhao \& Akoglu(2019)Zhao and Akoglu]{zhao2019pairnorm}
Lingxiao Zhao and Leman Akoglu.
\newblock Pairnorm: Tackling oversmoothing in gnns.
\newblock \emph{arXiv preprint arXiv:1909.12223}, 2019.

\bibitem[Zhao et~al.(2021)Zhao, Liu, Neves, Woodford, Jiang, and Shah]{zhao2021data}
Tong Zhao, Yozen Liu, Leonardo Neves, Oliver Woodford, Meng Jiang, and Neil Shah.
\newblock Data augmentation for graph neural networks.
\newblock In \emph{Proceedings of the aaai conference on artificial intelligence}, 2021.

\bibitem[Zhao et~al.(2022{\natexlab{a}})Zhao, Jin, Liu, Wang, Liu, Günneman, Shah, and Jiang]{zhao2022graph}
Tong Zhao, Wei Jin, Yozen Liu, Yingheng Wang, Gang Liu, Stephan Günneman, Neil Shah, and Meng Jiang.
\newblock Graph data augmentation for graph machine learning: A survey.
\newblock \emph{arXiv preprint arXiv:2202.08871}, 2022{\natexlab{a}}.

\bibitem[Zhao et~al.(2022{\natexlab{b}})Zhao, Liu, Wang, Yu, and Jiang]{zhao2022learning}
Tong Zhao, Gang Liu, Daheng Wang, Wenhao Yu, and Meng Jiang.
\newblock Learning from counterfactual links for link prediction.
\newblock In \emph{International Conference on Machine Learning}. PMLR, 2022{\natexlab{b}}.

\bibitem[Zheng et~al.(2021)Zheng, Huang, Rao, Katariya, Wang, and Subbian]{zheng2021cold}
Wenqing Zheng, Edward~W Huang, Nikhil Rao, Sumeet Katariya, Zhangyang Wang, and Karthik Subbian.
\newblock Cold brew: Distilling graph node representations with incomplete or missing neighborhoods.
\newblock \emph{arXiv preprint arXiv:2111.04840}, 2021.

\bibitem[Zhou \& Cao(2021)Zhou and Cao]{zhou2021overcoming}
Fan Zhou and Chengtai Cao.
\newblock Overcoming catastrophic forgetting in graph neural networks with experience replay.
\newblock In \emph{Proceedings of the AAAI conference on artificial intelligence}, volume~35, pp.\  4714--4722, 2021.

\bibitem[Zhu et~al.(2019)Zhu, Lin, He, Wang, Guan, Liu, and Cai]{zhu2019addressing}
Yu~Zhu, Jinghao Lin, Shibi He, Beidou Wang, Ziyu Guan, Haifeng Liu, and Deng Cai.
\newblock Addressing the item cold-start problem by attribute-driven active learning.
\newblock \emph{IEEE Transactions on Knowledge and Data Engineering}, 2019.

\bibitem[Zhu et~al.(2021)Zhu, Zhang, Xhonneux, and Tang]{zhu2021neural}
Zhaocheng Zhu, Zuobai Zhang, Louis-Pascal Xhonneux, and Jian Tang.
\newblock Neural bellman-ford networks: A general graph neural network framework for link prediction.
\newblock \emph{Advances in Neural Information Processing Systems}, 2021.

\end{thebibliography}
\bibliographystyle{tmlr}

\clearpage
\appendix

\part{}
\localtableofcontents
\clearpage

\section{Related Work}
\textbf{LP with GNNs.} Over the past few years, GNN architectures~\citep{kipf2016semi, gilmer2017neural, hamilton2017inductive, velivckovic2017graph, xu2018representation} have gained significant attention and demonstrated promising outcomes in LP tasks. There are two primary approaches to applying GNNs in LP. The first approach involves a node-wise encoder-decoder framework, which we discussed in \cref{sec:preliminary}. The second approach reformulates LP tasks as enclosing subgraph classification tasks~\citep{zhang2018link, cai2020multi, cai2021line, dong2022fakeedge}. Instead of directly predicting links, these methods perform graph classification tasks on the enclosing subgraphs sampled around the target link. These methods can achieve even better results compared to node-wise encoder-decoder frameworks by assigning node labels to indicate different roles within the subgraphs. However, constructing subgraphs poses challenges in terms of efficiency and scalability, requiring substantial computational resources. Our work focuses on the encoder-decoder framework for LP, circumventing the issues associated with subgraph construction.

\textbf{Methods for Cold-start Nodes.} Recently, several GNN-based methods~\citep{wu2019net, liu2020towards, tang2020investigating, liu2021tail, zheng2021cold} have explored degree-specific transformations to address robustness and cold-start node issues. Tang et al.\citep{tang2020investigating} introduced a degree-related graph convolutional network to mitigate degree-related bias in node classification tasks. Liu et al.\citep{liu2021tail} proposed a transferable neighborhood translation model to address missing neighbors for cold-start nodes. Zheng et al.\citep{zheng2021cold} tackled the cold-start nodes problem by recovering missing latent neighbor information. These methods require cold-start-node-specific architectural components, unlike our approach, which does not necessitate any architectural modifications. Additionally, other studies have focused on long-tail scenarios in various domains, such as cold-start recommendation\citep{chen2020esam, lu2020meta, hao2021pre}. Imbalance tasks present another common long-tail problem, where there are long-tail instances within small classes~\citep{lin2017focal, ren2020balanced, tan2020equalization, kang2019decoupling, tang2020long}. Approaches like~\citep{lin2017focal, ren2020balanced, tan2020equalization} address this issue by adapting the loss for different samples. However, due to the different problem settings, it is challenging to directly apply these methods to our tasks. We only incorporate the balanced cross entropy introduced by Lin et al.~\citep{lin2017focal} as one of our baselines. In addition to these one-shot training methods, many recommendation studies~\citep{kirkpatrick2017overcoming,xu2020graphsail,zhou2021overcoming,valkanas2024personalized} have addressed the issue of incorporating new nodes through incremental learning frameworks, where new nodes are continually added into the training process and specific strategies are designed to alleviate forgetting previous knowledge (i.e., warm nodes or old classes).

\textbf{Graph Data Augmentation.} Graph data augmentation expands the original data by perturbing or modifying the graphs to enhance the generalizability of GNNs~\citep{zhao2022graph, ding2022data}. Existing methods primarily focus on semi-supervised node-level tasks\citep{rong2019dropedge, feng2020graph, zhao2021data, park2021metropolis} and graph-level tasks~\citep{liu2022graph, luo2022automated}. However, the exploration of graph data augmentation for LP remains limited~\citep{zhao2022learning}. CFLP~\citep{zhao2022learning} proposes the creation of counterfactual links to learn representations from both observed and counterfactual links. Nevertheless, this method encounters scalability issues due to the high computational complexity associated with finding counterfactual links. Moreover, there exist general graph data augmentation methods~\citep{liu2022local, hu2022tuneup} that can be applied to various tasks. LAGNN~\citep{liu2022local} proposed to use a generative model to provide additional neighbor features for each node. TuneUP~\citep{hu2022tuneup} designs a two-stage training strategy, which trains GNNs twice to make them perform well on both warm nodes and cold-start nodes. These augmentation methods come with the trade-off of introducing extra runtime either before or during the model training. Unlike TLC-GNN~\citep{yan2021link}, which necessitates extracting topological features for each node pair, and GIANT~\citep{chien2021node}, which requires pre-training of the text encoder to improve node features, our methods are more streamlined and less complex.

\section{Additional Datasets Details}
\label{sec:data}

This section provides detailed information about the datasets used in our experiments. We consider various types of networks, including citation networks, collaboration networks, and co-purchase networks. The datasets we utilize are as follows:

\begin{itemize}[leftmargin=0.2in]
    \item Citation Networks: \cora and \citeseer originally introduced by \citet{yang2016revisiting}, consist of citation networks where the nodes represent papers and the edges represent citations between papers. \igbtiny~\citep{igbdatasets} is a recently-released benchmark citation network with high-quality node features and a large dataset size. 
    \item Collaboration Networks: \cs and \physics are representative collaboration networks. In these networks, the nodes correspond to authors and the edges represent collaborations between authors. 
    \item Co-purchase Networks: \computers and \photos are co-purchase networks, where the nodes represent products and the edges indicate the co-purchase relationship between two products.
\end{itemize}

\textbf{Why there are no OGB~\citep{hu2020open} datasets applied?} OGB benchmarks that come with node features, such as OGB-collab and OGB-citation2, lack a substantial number of isolated or low-degree nodes, which makes it challenging to yield convincing results for experiments focusing on the cold-start problem. This is primarily due to the split setting adopted by OGB, where the evaluation is centered around a set of the most recent papers with high degrees. Besides, considering these datasets have their fixed splitting settings based on time, it will lead to inconsistent problems to compared with the leaderboard results if 
 we use our own splitting method to ensure we have a reasonable number of isolated/low-degree nodes. Given these constraints, we opted for another extensive benchmark dataset, IGB-100K~\citep{igbdatasets}, to test and showcase the effectiveness of our methods on large-scale graphs. We further conducted the experiments on \igbsmall, which are shown in~\cref{sec:largescale}.

\begin{table}
\centering
\small
\caption{Detailed statistics of data splits under the transductive and inductive setting. }
\label{tab:data}
\scalebox{0.93}{
\begin{tabular}{l|cc||cc|cc|cc} 
\toprule
\rowcolor{gray!20} \multicolumn{9}{c}{\textbf{Transductive Setting}}\\\midrule
\multirow{2}{*}{Datasets} & \multicolumn{2}{c||}{Original Graph} & \multicolumn{2}{c|}{Testing \isolated} &\multicolumn{2}{c|}{Testing \cold} &\multicolumn{2}{c}{Testing \warm}\\
&\#Nodes &\#Edges &\#Nodes &\#Edges &\#Nodes &\#Edges &\#Nodes &\#Edges\\\midrule
\cora & 2,708  & 5,278   & 135 & 164 & 541  & 726  & 662   & 1,220 \\
\citeseer &3,327  & 4,552 & 291 & 342 & 492  & 591  & 469  & 887\\
\cs &18,333 & 163,788 & 309 & 409 & 1,855 & 2,687 & 10,785 & 29,660  \\
\physics &34,493 & 495,924 & 275 & 397 & 2,062 & 3,188 & 25,730 & 95,599\\
\computers & 13,752 & 491,722 & 218 & 367 & 830  & 1,996 & 11,887 & 194,325 \\
\photos& 7,650  & 238,162 & 127 & 213 & 516  & 1,178 & 6,595  & 93,873 \\
\igbtiny&100,000 &547,416 &1,556 & 1,737 & 6,750 & 7,894 & 23,949 &35,109\\\midrule
\rowcolor{gray!20} \multicolumn{9}{c}{\textbf{Inductive Setting}}\\\midrule
\multirow{2}{*}{Datasets} & \multicolumn{2}{c||}{Original Graph} & \multicolumn{2}{c|}{Testing \isolated} &\multicolumn{2}{c|}{Testing \cold} &\multicolumn{2}{c}{Testing \warm}\\
&\#Nodes &\#Edges &\#Nodes &\#Edges &\#Nodes &\#Edges &\#Nodes &\#Edges\\\midrule
\cora & 2,708  & 5,278 &149 &198 &305 &351 &333 & 505 \\
\citeseer &3,327  & 4,552 &239 &265 &272 & 302 &239 &339 \\
\cs &18,333 & 163,788 &1,145 &1,867 &1,202 &1,476 &6,933 &13,033  \\
\physics &34,493 & 495,924 &2,363 &5,263 &1,403 &1,779 & 17,881 &42,548 \\
\computers & 13,752 & 491,722 &1,126 &4,938 &239 &302 &9,235 &43,928 \\
\photos& 7,650  & 238,162 &610 &2,375 &169 &212 &5,118 &21,225 \\
\igbtiny&100,000 &547,416 &5,507 &9,708 &8,706 &13,815 &24,903 &41,217\\
\bottomrule
\end{tabular}
}
\end{table}

\subsection{Transductive Setting}
For the transductive setting, we randomly split the edges into training, validation, and testing sets based on the splitting ratio specified in \cref{sec:experimental_settings}. The nodes in training/validation/testing are all visible during the training process. However, the positive edges in validation/testing sets are masked out for training. After the split, we calculate the degrees of each node using the validation graph. The dataset statistics are shown in \cref{tab:data}. 

\subsection{Inductive Setting}
The inductive setting is considered a more realistic setting compared to the transductive setting, where new nodes appear after the training process. Following the inductive setting introduced in \citet{guo2022linkless} and \citet{shiao2022link}, we perform node splitting to randomly sample 10\% nodes from the original graph as the new nodes appear after the training process. The remaining nodes are considered observed nodes during the training. Next, we group the edges into three sets: observed-observed, observed-new, and new-new node pairs. We select 10\% of observed-observed, 10\% of observed-new, and 10\% of new-new node pairs as the testing edges. We consider the remaining observed-new and new-new node pairs, along with an additional 10\% of observed-observed node pairs, as the newly visible edges for the testing inference. The datasets statistics are shown in \cref{tab:data}.

\section{Further Experimental Results}
\label{sec:further_results}

\subsection{Selection of the threshold $\delta$.}
\label{sec:preanalysis_appendix}
\begin{figure}
    \centering
    \subfigure[\cora]
    {\includegraphics[width=0.44\linewidth]{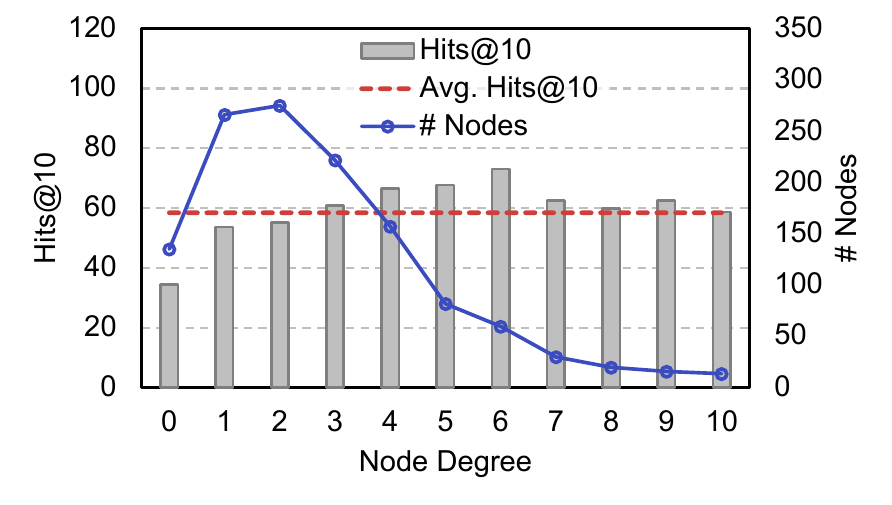}}\label{fig:preanalysis_cora}
    \subfigure[\cs]
    {\includegraphics[width=0.44\linewidth]{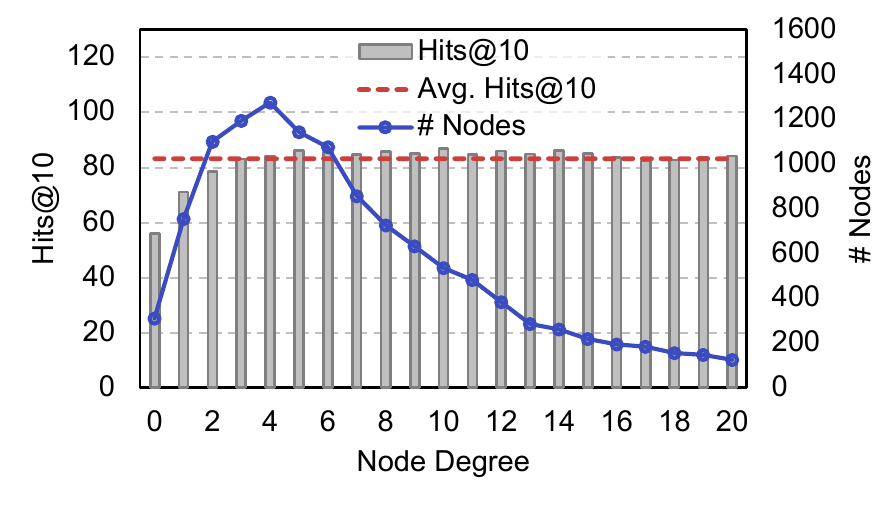}}\label{fig:preanalysis_cs}
    \subfigure[\computers]
    {\includegraphics[width=0.44\linewidth]{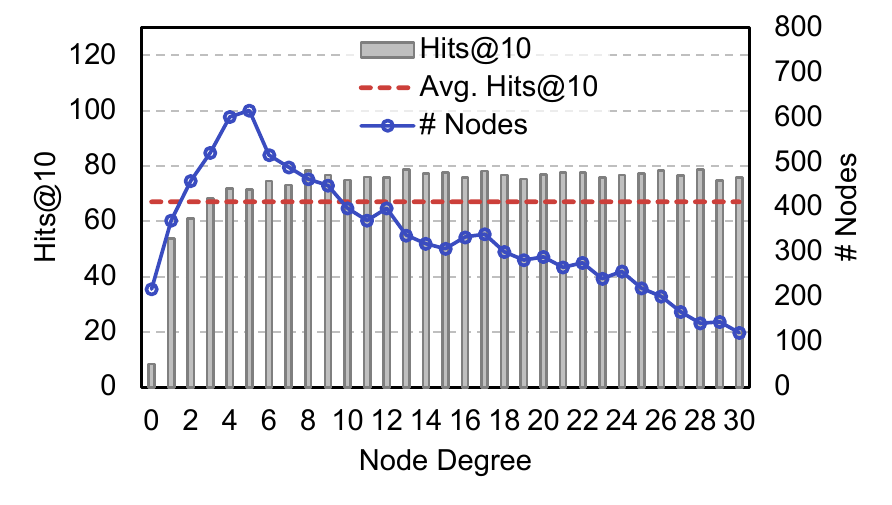}}\label{fig:preanalysis_computers}
    \subfigure[\igbtiny]
    {\includegraphics[width=0.44\linewidth]{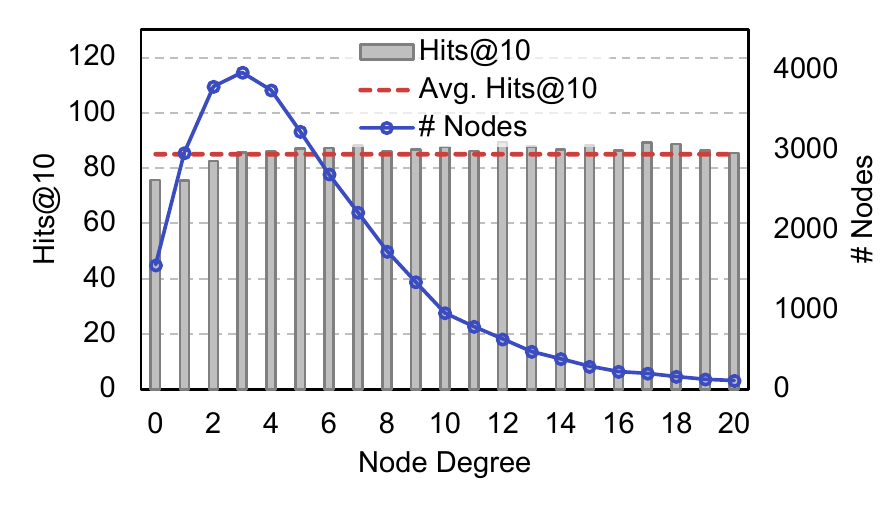}}\label{fig:preanalysis_igb}
    \caption{Node Degree Distribution and LP Performance Distribution w.r.t Nodes Degrees showing reverse trends on various datasets.}
    \label{fig:pre-analysis-appendix}
\end{figure}

Our decision to set the threshold $\delta$ at 2 is grounded in data-driven analysis, as illustrated in \cref{fig:pre-analysis} and \cref{fig:pre-analysis-appendix}. These figures reveal that nodes with degrees not exceeding 2 consistently perform below the average Hits@10 across all datasets, and higher than 2 will outperform the average results. Besides, our choice aligns with methodologies in previous studies~\citep{liu2020towards, liu2021tail}, where cold nodes are identified using a fixed threshold across all the datasets. In addition, we conduct experiments with different thresholds $\delta$ on \cora and \citeseer datasets. The results are shown in~\cref{tab:threshold}. Our findings were consistent across different thresholds, with similar observations at $\delta$ = 1, $\delta$ = 2, and $\delta$ = 3. This indicates that our method's effectiveness is not significantly impacted by changes in this threshold.

\begin{table}
\centering
\caption{Performance with different thresholds $\delta$ on \cora and \citeseer datasets.}
\label{tab:threshold}
\scalebox{0.8}{
\begin{tabular}{l|c||c|c||c|c||c|c}
\toprule
         &            & \multicolumn{2}{c||}{$\delta$ = 1}        & \multicolumn{2}{c||}{$\delta$ = 2}        & \multicolumn{2}{c}{$\delta$ = 3}        \\\midrule
         &            & Gsage        & \ours & Gsage        & \ours & Gsage        & \ours \\\midrule
\multirow{4}{*}{\cora}     & Isolated   & \ms{31.34}{5.60} & \ms{ 42.20}{2.30} & \ms{ 32.20}{3.58} & \ms{44.27}{3.82} & \ms{ 31.95}{1.26} & \ms{43.17}{2.94}  \\
         & Low-degree & \ms{53.98}{1.20} & \ms{57.99}{1.34} & \ms{ 59.45}{1.09} & \ms{61.98}{1.14} & \ms{ 59.64}{1.01} & \ms{62.68}{0.63}  \\
         & Warm      & \ms{ 61.68}{0.29} & \ms{61.17}{0.43} & \ms{ 61.14}{0.78} & \ms{59.07}{0.68} & \ms{ 61.03}{0.79} & \ms{59.91}{0.44}  \\
         & Overall  & \ms{ 58.01}{0.57} & \ms{59.16}{0.44} & \ms{ 58.31}{0.68} & \ms{58.92}{0.82} & \ms{ 58.08}{0.74} & \ms{59.99}{0.53}  \\\midrule
\multirow{4}{*}{\citeseer} & Isolated   & \ms{47.25}{1.82} & \ms{56.49}{1.72} & \ms{ 47.13}{2.43} & \ms{57.54}{1.04} & \ms{ 47.31}{2.17} & \ms{56.90}{1.12}  \\
         & Low-degree & \ms{54.10}{0.85} & \ms{71.09}{0.47} & \ms{ 61.88}{0.79} & \ms{75.50}{0.39} & \ms{ 62.97}{0.83} & \ms{75.45}{0.40}  \\
         & Warm       & \ms{72.41}{0.35} & \ms{74.57}{1.04} & \ms{ 71.45}{0.52} & \ms{74.68}{0.67} & \ms{ 73.57}{0.46} & \ms{75.02}{0.84}  \\
         & Overall    & \ms{64.27}{0.45} & \ms{70.53}{0.91} & \ms{ 63.77}{0.83} & \ms{71.73}{0.47} & \ms{ 64.05}{0.42} & \ms{71.80}{0.40} \\
\bottomrule
\end{tabular}}
\end{table}

\subsection{Further analysis about the duplication node types}
\label{sec:nodetypes}

To make our analysis about the duplication node types more comprehensive, we conducted additional experiments to evaluate different node duplication strategies under both the \ours and \ourslight settings. We present detailed experiments on the Citeseer dataset, where we duplicate different subsets of nodes: isolated nodes, cold nodes, mid-warm nodes, warm nodes, random nodes, and all nodes. We report the corresponding training time, memory usage, and performance across multiple evaluation subsets. The results are summarized in~\cref{tab:ablation_duplicatednodes_appendix}.

From the table, we can observe that for both \ours and \ourslight, duplicating mid-warm and warm nodes results in little to no performance improvement. Random duplication provides clear improvements over both no duplication and warm node duplication, but remains less effective than cold node duplication. Since random duplication and cold nodes duplication duplicate the same number of nodes, they incur similar memory and training costs. Under the \ours setting, compared to cold nodes duplication, all nodes duplication increases training time by approximately 32\% (3.3s vs. 2.5s) and memory usage by approximately 35\% (93.99MB vs. 69.46MB), while achieving only marginal additional performance gain (Low-degree Hits@10: 76.09 vs. 75.50). Under the \ourslight setting, full duplication increases training time by approximately 14\% (2.4s vs. 2.1s), while memory usage remains identical, again with limited additional performance gain (Low-degree Hits@10: 75.32 vs. 74.99).

These results demonstrate that our proposed selective cold node duplication achieves most of the performance improvement while significantly reducing both training time and memory cost compared to full duplication. This further supports the practical advantage of our method under constrained compute budgets.

\begin{table}[]
\centering
\vspace{-0.2in}
\caption{Computation time, memory usage, and link prediction performance with different duplication nodes of \ours and \ourslight on Citeseer. "D\_*" indicates duplication of "*" group nodes for one time.}
\label{tab:ablation_duplicatednodes_appendix}
\scalebox{0.85}{
\begin{tabular}{l|c|cc|cccc}
\toprule
\rowcolor{gray!20} \multicolumn{8}{c}{\textbf{\ours}}\\ \midrule
\multicolumn{1}{c|}{} & Time(s)  & \multicolumn{2}{c|}{Memory(MB)} & \multicolumn{4}{c}{Performance   (Hits@10)} \\ \midrule
\multicolumn{1}{c|}{} & Training & Node attributes     & Edges    & Isolated   & Low-degree  & Warm  & Overall  \\ \midrule
Supervised  & 1.8 & 47.00    & 0.14 & 47.13 & 61.88 & 71.45 & 63.77 \\
D\_Isolated & 1.9 & 54.03 & 0.14 & 54.04 & 72.28 & 74.53 & 69.95 \\
D\_Cold     & 2.4 & 69.46 & 0.14 & 57.54 & 75.50  & 74.68 & 71.73 \\
D\_Mid-warm & 2.2 & 55.83 & 0.14 & 46.93 & 61.34 & 71.84 & 63.75 \\
D\_Warm     & 2.2 & 57.53 & 0.14 & 47.49 & 62.20  & 71.54 & 63.99 \\
D\_Random   & 2.4 & 69.46 & 0.14 & 54.10  & 72.39 & 75.05 & 70.06 \\
D\_All      & 3.3 & 93.99 & 0.14 & 58.87 & 76.09 & 76.01 & 72.44 \\\midrule
\rowcolor{gray!20} \multicolumn{8}{c}{\textbf{\ourslight}}                                   \\ \midrule
                     & Time(s)  & \multicolumn{2}{c|}{Memory(MB)} & \multicolumn{4}{c}{Performance(Hits@10)}    \\ \midrule
                     & Training & Node attributes     & Edges    & Isolated   & Low-degree  & Warm  & Overall  \\ \midrule
Supervised  & 1.8 & 47.00    & 0.14 & 47.13 & 61.88 & 71.45 & 63.77 \\
D\_Isolated & 2.0   & 47.00    & 0.14 & 49.06 & 69.95 & 74.67 & 68.32 \\
D\_Cold     & 2.1 & 47.00    & 0.14 & 52.46 & 73.71 & 74.99 & 70.34 \\
D\_Mid-warm & 2.2 & 47.00    & 0.14 & 46.05 & 61.57 & 72.30  & 63.88 \\
D\_Warm     & 2.1 & 47.00    & 0.14 & 46.53 & 61.86 & 71.77 & 63.82 \\
D\_Random   & 2.2 & 47.00    & 0.14 & 48.92 & 69.59 & 73.90  & 67.81 \\
D\_All      & 2.4 & 47.00    & 0.14 & 52.44 & 74.05 & 75.21 & 70.49 \\
\bottomrule
\end{tabular}}
\end{table}

\begin{wraptable}{r}{0.5\linewidth}
\vspace{-0.3in}
\caption{Performance on the large-scale dataset. The best result is \textbf{bold}. Our method consistently outperforms GSage on \igbsmall.}
\label{tab:largescale}
\scalebox{0.9}{
\begin{tabular}{l|c|c|c}
\toprule
       &            & GSage        & NodeDup               \\\midrule
\multirow{4}{*}{\igbsmall} & Isolated   & \ms{82.10}{0.06} & \textbf{\ms{87.81}{0.40}} \\
       & Low-degree & \ms{84.73}{0.06} & \textbf{\ms{90.84}{0.03}} \\
       & Warm       & \ms{89.98}{0.02} & \textbf{\ms{91.31}{0.02}} \\
       & Overall    & \ms{89.80}{0.02} & \textbf{\ms{91.29}{0.02}} \\
\bottomrule
\end{tabular}}
\vspace{-0.1in}
\end{wraptable}

\subsection{Performance on large-scale datasets}
\label{sec:largescale}
As outlined in~\cref{sec:proposed_method}, our methods incur a minimal increase in time complexity compared to base GNNs, with the increase being linearly proportional to the number of cold nodes. This ensures scalability. Besides, the effectiveness of our method is also insensitive to dataset size. We extend our experiments to the \igbsmall dataset, featuring 1 million nodes and 12 million edges. The findings, which we detail in~\cref{tab:largescale}, affirm the effectiveness of our methods in handling large-scale datasets, consistent with observations from smaller datasets. 

\begin{table}[]
\centering
\small
\caption{Performance on heterophilic datasets. The best result for each dataset is \textbf{bold}.}
\label{tab:heterophilic}
\scalebox{0.9}{
\begin{tabular}{lllll}
\toprule
          &            & GSage        & NodeDup(L)            & NodeDup               \\ \midrule
\multirow{4}{*}{Chameleon} & Isolated   & \ms{24.91}{6.75} & \textbf{\ms{30.76}{4.02}} & \ms{27.37}{2.88}          \\
          & Low-degree & \ms{79.09}{1.21} & \ms{80.11}{0.68}          & \textbf{\ms{80.91}{0.41}} \\
          & Warm       & \ms{94.00}{0.23} & \textbf{\ms{94.01}{0.12}} & \ms{93.68}{0.44}          \\
          & Overall    & \ms{92.77}{0.19} & \textbf{\ms{92.88}{0.10}} & \ms{92.57}{0.44}          \\ \midrule
\multirow{4}{*}{Squirrel}  & Isolated   & \ms{25.05}{3.70} & \textbf{\ms{33.07}{3.20}} & \ms{30.11}{1.57}          \\
          & Low-degree & \ms{63.34}{2.12} & \ms{66.61}{0.26}          & \textbf{\ms{68.05}{0.80}} \\
          & Warm       & \ms{93.35}{0.22} & \ms{93.43}{0.11}          & \textbf{\ms{93.82}{0.13}} \\
          & Overall    & \ms{92.89}{0.23} & \ms{93.02}{0.11}          & \textbf{\ms{93.41}{0.13}} \\
\bottomrule
\end{tabular}}
\end{table}

\subsection{Performance on heterophily datasets}
\label{sec:heterophily}

We have conducted experiments on two heterophilic datasets (i.e., Chameleon~\citep{pei2020geom} and Squirrel~\citep{pei2020geom}), with the results shown in~\cref{tab:heterophilic}. Our methods improve GNN performance across all settings on these datasets. Specifically, NodeDup and NodeDup(L) enhance the performance of \isolated nodes by 9.9\% and 23.5\% on Chameleon, and by 20.2\% and 32.0\% on Squirrel.

\begin{wraptable}{r}{0.45\linewidth}
\vspace{-0.3in}
\caption{Performance on the MovieLens. The best result is \textbf{bold}. Our method consistently outperforms GSage.}
\label{tab:recsys}
\scalebox{0.9}{
\begin{tabular}{lccc}
\toprule
\multicolumn{1}{c}{} & \multicolumn{3}{c}{MovieLens\_1M} \\ \midrule
\multicolumn{1}{c}{} & GSage    & NodeDup(L)   & NodeDup  \\ \midrule
Isolated             & 0       & 3.08         & \textbf{5.38}     \\
Low\_degree          & 30.7    & 35.07        & \textbf{37.69}    \\
Warm                 & 41.79   & 44.64        & \textbf{45.52}    \\
Overall              & 41.71   & 44.56        & \textbf{45.78}\\
\bottomrule
\end{tabular}}
\vspace{-0.1in}
\end{wraptable}

\subsection{Performance on recommendation datasets}
\label{sec:recsys}
To further evaluate the practical applicability of our method to real-world recommendation scenarios, we followed the approach introduced in KGCN~\citep{huang2021knowledge} to construct a movie-movie graph, where two movies are connected if they received high ratings from the same user. This graph is then used to recommend similar movies to users. This setup forms an item-based collaborative filtering recommendation task, allowing us to apply our methods. The results is shown in~\cref{tab:recsys}. Compared to the baseline, both \ours and \ourslight achieve consistent improvements, particularly on low-degree and isolated nodes, where these cold-start items often limits recommendation accuracy.

\subsection{Performance compared with heuristic methods}
\label{sec:heuristic}
We compare our method with traditional link prediction baselines, such as common neighbors (CN), Adamic-Adar(AA), Resource allocation (RA). The results are shown in~\cref{tab:heuristic}. We observe that \ours can consistently outperform these heuristic methods across all the datasets, with particularly significant improvements observed on \isolated nodes. 

\begin{table}
\centering
\vspace{-0.2in}
\caption{Performance compared with heuristic methods, Upsampling and DegFairGNN~\citep{liu2023generalized}. The best result is \textbf{bold}.}
\label{tab:heuristic}
\scalebox{0.8}{
\begin{tabular}{l|c||c|c|c||c|c|c||c}
\toprule
          &            & CN & AA & RA &Upsampling &DegFairGNN & GSage        & \ours     \\\midrule
\multirow{4}{*}{\cora}      & Isolated   & 0.00        & 0.00        & 0.00  &\ms{32.81}{2.75}  &\ms{18.70}{1.53}     & \ms{32.20}{3.58}          & \textbf{\ms{44.27}{3.82}}\\
          & Low-degree & 20.30       & 20.14       & 20.14   &\ms{59.57}{0.60} &\ms{38.43}{0.14}   & \ms{59.45}{1.09}          & \textbf{\ms{61.98}{1.14}}\\
          & Warm       & 38.33       & 38.90       & 38.90   &\ms{60.49}{0.81} &\ms{42.49}{1.82}    & \textbf{\ms{61.14}{0.78}} & \ms{59.07}{0.68}          \\
          & Overall    & 25.27       & 25.49       & 25.49    &\ms{57.90}{0.65} &\ms{39.24}{1.10}   & \ms{58.31}{0.68}          & \textbf{\ms{58.92}{0.82}}\\\midrule
\multirow{4}{*}{\citeseer}  & Isolated   & 0.00        & 0.00        & 0.00 &\ms{46.88}{0.45}    &\ms{15.50}{1.27}    & \ms{47.13}{2.43}          & \textbf{\ms{57.54}{1.04}}\\
          & Low-degree & 26.86       & 27.00       & 27.00   &\ms{62.32}{1.57} &\ms{45.06}{0.96}    & \ms{61.88}{0.79}          & \textbf{\ms{75.50}{0.39}}\\
          & Warm       & 37.30       & 39.02       & 39.02   &\ms{71.33}{1.35} &\ms{55.47}{1.08}    & \ms{71.45}{0.52}          & \textbf{\ms{74.68}{0.67}}\\
          & Overall    & 30.81       & 31.85       & 31.85    &\ms{63.81}{0.81} &\ms{44.58}{1.03}   & \ms{63.77}{0.83}          & \textbf{\ms{71.73}{0.47}}\\\midrule
\multirow{4}{*}{\cs}        & Isolated   & 0.00        & 0.00        & 0.00 &\ms{49.63}{2.24}    &\ms{17.93}{1.35}    & \ms{56.41}{1.61}          & \textbf{\ms{65.87}{1.70}}\\
          & Low-degree & 39.60       & 39.60       & 39.60    &\ms{75.62}{0.13} &\ms{49.83}{0.68}   & \ms{75.95}{0.25}          & \textbf{\ms{81.12}{0.36}}\\
          & Warm       & 72.73       & 72.74       & 72.72    &\ms{83.40}{0.73} &\ms{61.72}{0.37}   & \ms{84.37}{0.46}          & \textbf{\ms{84.76}{0.41}}\\
          & Overall    & 69.10       & 69.11       & 69.10    &\ms{82.34}{0.64} &\ms{60.20}{0.37}   & \ms{83.33}{0.42}          & \textbf{\ms{84.23}{0.39}}\\\midrule
\multirow{4}{*}{\physics}   & Isolated   & 0.00        & 0.00        & 0.00  &\ms{52.01}{0.97}  &\ms{19.48}{2.94}     & \ms{47.41}{1.38}          & \textbf{\ms{66.65}{0.95}}\\
          & Low-degree & 46.08       & 46.08       & 46.08   &\ms{79.63}{0.13} &\ms{47.63}{0.52}    & \ms{79.31}{0.28}          & \textbf{\ms{84.04}{0.22}}\\
          & Warm       & 85.48       & 85.74       & 85.70    &\ms{89.41}{0.32} &\ms{62.79}{0.82}   & \ms{90.28}{0.23}          & \textbf{\ms{90.33}{0.05}}\\
          & Overall    & 83.87       & 84.12       & 84.09    &\ms{89.33}{0.46} &\ms{62.13}{0.76}   & \ms{89.76}{0.22}          & \textbf{\ms{90.03}{0.05}}\\\midrule
\multirow{4}{*}{\computers} & Isolated   & 0.00        & 0.00        & 0.00 &\ms{11.36}{0.72}  &\ms{9.36}{1.81}     & \ms{9.32}{1.44}           & \textbf{\ms{19.62}{2.63}}\\
          & Low-degree & 28.31       & 28.31       & 28.31   &\ms{58.23}{0.88} &\ms{18.90}{0.81}    & \ms{57.91}{0.97}          & \textbf{\ms{61.16}{0.92}}\\
          & Warm       & 59.67       & 63.50       & 62.84   &\ms{67.07}{0.49} &\ms{31.44}{2.25}    & \ms{66.87}{0.47}          & \textbf{\ms{68.10}{0.25}}\\
          & Overall    & 59.24       & 63.03       & 62.37   &\ms{66.87}{0.48} &\ms{31.27}{2.22}    & \ms{66.67}{0.47}          & \textbf{\ms{67.94}{0.25}}\\\midrule
\multirow{4}{*}{\photos}    & Isolated   & 0.00        & 0.00        & 0.00 &\ms{10.92}{2.15}  &\ms{12.99}{1.51}     & \ms{9.25}{2.31}           & \textbf{\ms{17.84}{3.53}}\\
          & Low-degree & 28.44       & 28.78       & 28.78   &\ms{51.67}{0.98} &\ms{20.18}{0.21}    & \ms{52.61}{0.88}          & \textbf{\ms{54.13}{1.58}}\\
          & Warm       & 64.53       & 67.26       & 66.88   &\ms{65.75}{0.73} &\ms{42.72}{0.89}    & \ms{67.64}{0.55}          & \textbf{\ms{68.68}{0.49}}\\
          & Overall    & 63.94       & 66.64       & 66.26   &\ms{65.45}{0.71} &\ms{42.37}{0.87}    & \ms{67.32}{0.54}          & \textbf{\ms{68.39}{0.48}}\\\midrule
\multirow{4}{*}{\igbtiny}  & Isolated   & 0.00        & 0.00        & 0.00  &\ms{75.49}{0.90}  &\ms{57.09}{21.08}    & \ms{75.92}{0.52}          & \textbf{\ms{88.04}{0.20}}\\
          & Low-degree & 12.26       & 12.26       & 12.26   &\ms{79.47}{0.11} &\ms{59.45}{21.84}    & \ms{79.38}{0.23}          & \textbf{\ms{88.98}{0.17}}\\
          & Warm       & 30.65       & 30.65       & 30.65   &\ms{86.54}{0.19} &\ms{65.57}{20.43}    & \ms{86.42}{0.24}          & \textbf{\ms{88.28}{0.20}}\\
          & Overall    & 26.22       & 26.22       & 26.22   &\ms{84.87}{0.14} &\ms{64.16}{20.70}    & \ms{84.77}{0.21}          & \textbf{\ms{88.39}{0.18}}\\
\bottomrule
\end{tabular}}
\vspace{-0.2in}
\end{table}

\subsection{Performance compared with additional cold-start methods}
\label{sec:appendix-cold}

\textbf{Upsampling~\citep{provost2000machine}.} In \cref{sec:method}, we discussed the issue of under-representation of cold nodes during the training of LP, which is the main cause of their unsatisfactory performance. To tackle this problem, one straightforward and naive approach is upsampling~\citep{provost2000machine}, which involves increasing the number of samples in the minority class. In order to further demonstrate the effectiveness of our methods, we conducted experiments where we doubled the edge sampling probability of code nodes, aiming to enhance their visibility. The results are presented in \cref{tab:heuristic}. We can observe that \ours outperforms upsampling in almost all the cases, except for \warm nodes on \cora.

\begin{table}
\centering
\caption{Performance compared with GRADE~\citep{wang2022uncovering} and SAILOR~\citep{liao2023sailor}. The best result is \textbf{bold}. }
\label{tab:additional_coldstart}
\scalebox{0.8}{
\begin{tabular}{l|c|cccccc} 
\toprule
         &            & GCN                   & GRADE        & SAILOR       & \ourslight        & \ours        \\ \midrule
\multirow{4}{*}{\cora}     & Isolated   & \ms{40.61}{3.52}          & \ms{43.29}{2.62} & \ms{45.12}{1.29} & \ms{42.93}{2.68}          & \textbf{\ms{46.71}{1.53}} \\
         & Low-degree & \ms{63.86}{0.78}          & \ms{58.76}{1.27} & \ms{62.98}{3.92} & \textbf{\ms{64.63}{1.60}} & \ms{64.10}{1.37}          \\
         & Warm       & \ms{60.59}{0.62}          & \ms{60.00}{0.51} & \ms{57.34}{3.80} & \textbf{\ms{61.31}{0.43}} & \ms{60.26}{0.70}          \\
         & Overall    & \ms{60.16}{0.44}          & \ms{56.90}{0.71} & \ms{58.33}{3.51} & \textbf{\ms{61.02}{0.61}} & \ms{59.90}{0.89}          \\ \midrule
\multirow{4}{*}{\citeseer} & Isolated   & \ms{45.56}{1.30}          & \ms{50.11}{2.24} & \ms{49.29}{2.75} & \ms{47.84}{0.94}          & \textbf{\ms{50.64}{1.10}} \\
         & Low-degree & \ms{69.37}{0.36}          & \ms{59.49}{1.13} & \ms{65.78}{1.11} & \ms{70.15}{1.56}          & \textbf{\ms{71.13}{0.64}} \\
         & Warm       & \textbf{\ms{74.68}{0.38}} & \ms{70.01}{0.50} & \ms{72.66}{0.37} & \ms{73.26}{0.97}          & \ms{72.93}{0.78}          \\
         & Overall    & \ms{67.48}{0.42}          & \ms{61.11}{0.72} & \ms{64.80}{0.66} & \ms{67.47}{0.83}          & \textbf{\ms{67.67}{0.66}}\\
\bottomrule
\end{tabular}}
\end{table}
\textbf{The methods tackling degree bias in GNNs.} SAILOR~\citep{liao2023sailor} proposes a structural augmentation framework to enhance the representation learning of tail nodes. GRADE~\citep{wang2022uncovering} improves structural fairness using graph contrastive learning methods. We used GCN as the encoder for both \ourslight and \ours to ensure consistency, as both GRADE and SAILOR used GCN as their encoder. \cref{tab:additional_coldstart} shows that our methods outperform these baselines in all settings. Additionally, both GRADE and SAILOR perform better than vanilla GCN on \isolated nodes, which is the primary focus of their training. The sub-optimal performance of GRADE and SAILOR on Low-degree nodes in link prediction likely stems from the fact that their augmentation strategies, which are tailored for node classification, are less suited for ranking-based link prediction tasks. The specific reasons are as follows: GRADE encourages representation smoothness via degree-aware contrastive learning, which benefits node classification but reduces the embedding discrimination required for link prediction, particularly when ranking edges involving low-degree nodes, as reflected by Hits@10. SAILOR constructs pseudo-homophilic edges based on node labels; however, label similarity does not always align with link formation in link prediction, and adding label-based edges may introduce noise.

DegFairGNN~\citep{liu2023generalized} introduces a learnable debiasing function in the GNN architecture to produce fair representations for nodes, aiming for similar predictions for nodes within the same class, regardless of their degrees. Unfortunately, we've found in~\cref{tab:heuristic} that this approach is not well-suited for link prediction tasks for several reasons: (1) This method is designed specifically for node classification tasks. For example, the fairness loss, which ensures prediction distribution uniformity among low and high-degree node groups, is not suitable for link prediction because there is no defined node class in link prediction tasks. (2) This approach achieves significant performance in node classification tasks by effectively mitigating degree bias. However, in the context of link prediction, the degree trait is crucial. Applying DegFairGNN~\citep{liu2023generalized} would compromise the model's ability to learn from structural information, such as isomorphism and common neighbors. This, in turn, would negatively impact link prediction performance, as evidenced by references~\citep{zhang2018link, chamberlain2022graph}.

\begin{figure}[h]
    \centering
    \includegraphics[width=0.38\linewidth]{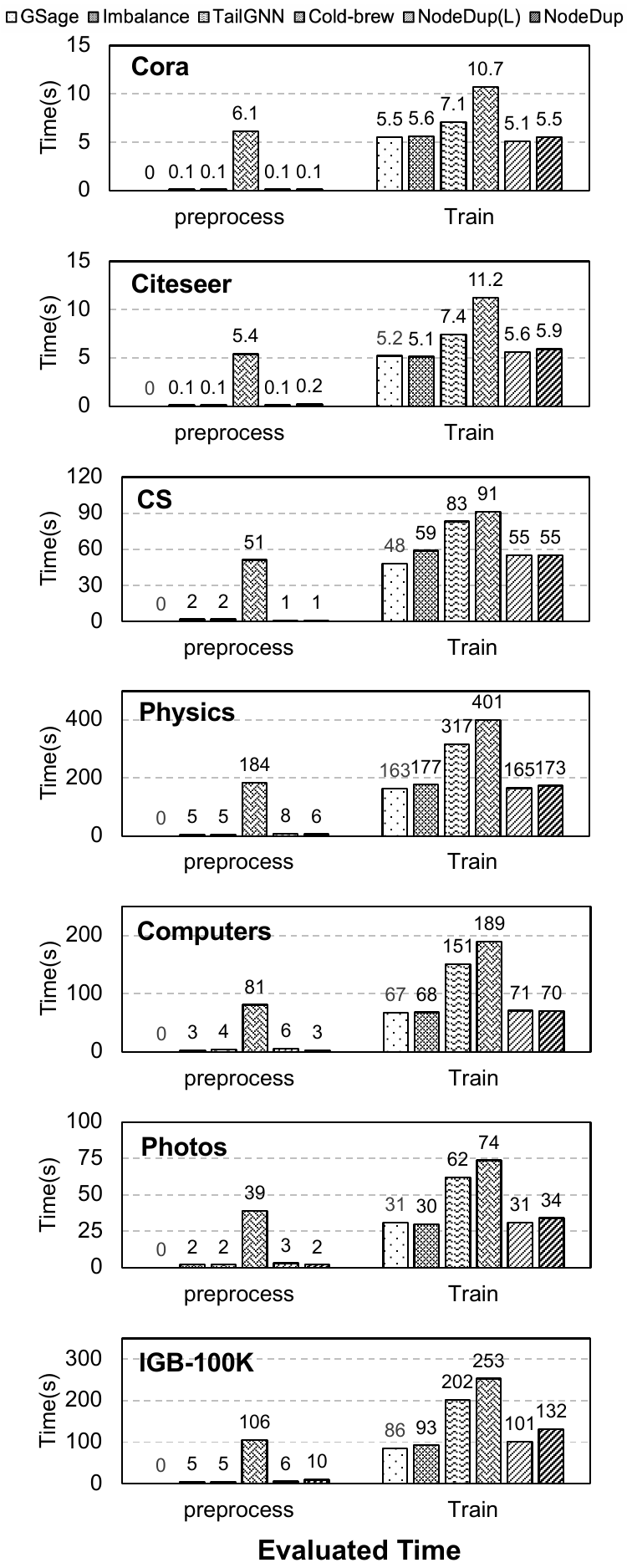}
    \caption{Time-consuming compared with cold-start methods. The histograms show the preprocessing and training time consumption of each method.}
    \label{fig:cold_time}
\end{figure}

\subsection{Efficiency comparison with the base GNN model and cold-start baselines}
\label{sec:appendix_coldtime}
The efficiency comparison between our methods and cold-start baselines is presented in~\cref{fig:cold_time}. We can observe that our methods and Imbalance exhibit similar efficiency, comparable to GSage. However, TailGNN and Cold-brew demand significantly more preprocessing and training time. Cold-brew, in particular, needs the most preprocessing time as it needs to train a teacher model for distillation. 

\subsection{Additional results compared with augmentation baselines}
\label{sec:append_aug}

\begin{figure}[t]
    \centering
    \includegraphics[width=\linewidth]{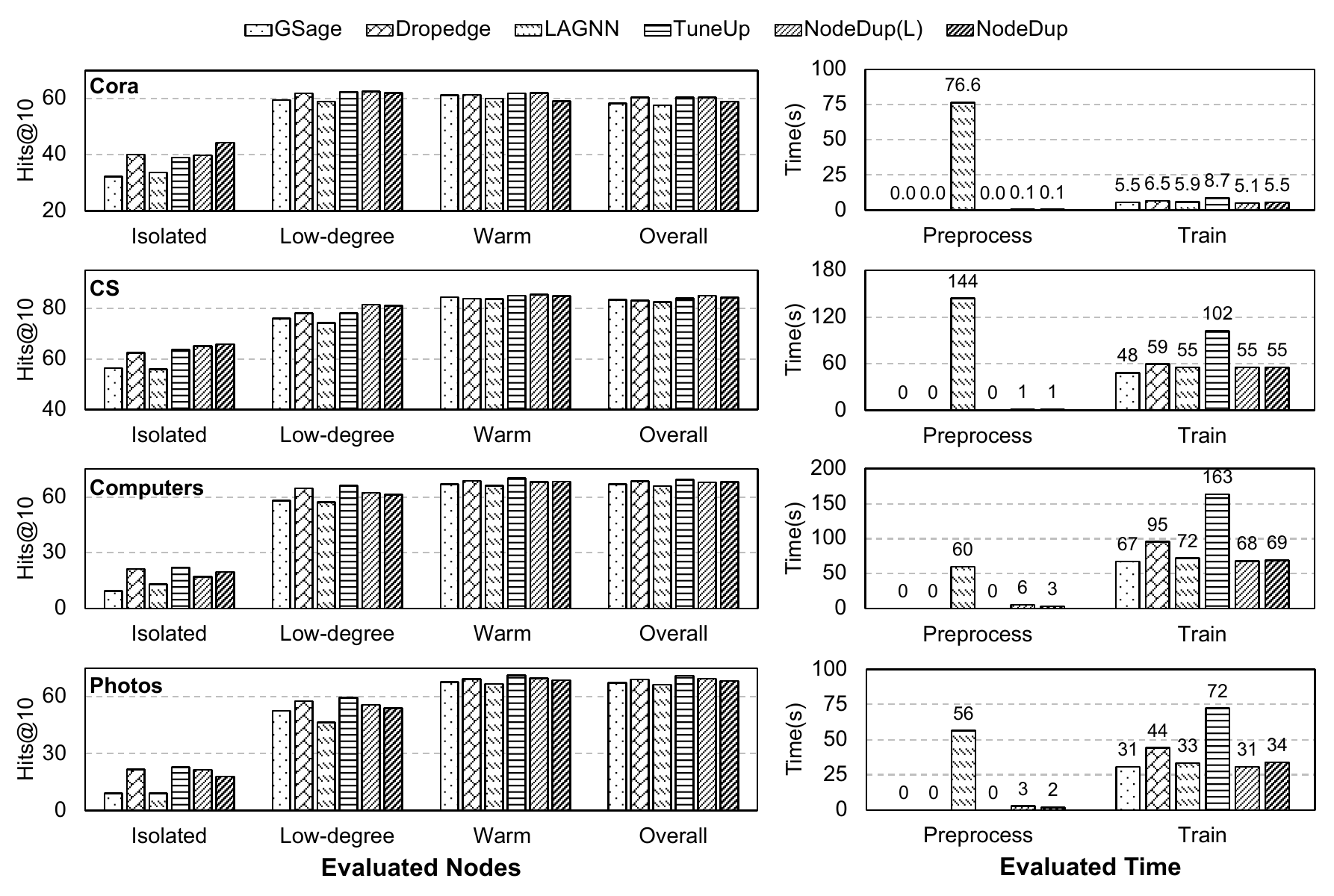}
    \caption{Performance and time-consuming compared with augmentation methods (Remaining results of \cref{fig:aug}). The \textit{left} histograms show the performance results, and the \textit{right} histograms show the preprocessing and training time consumption of each method.}
    \label{fig:aug_appendix}
\end{figure}

\cref{fig:aug_appendix} presents the performance compared with augmentation methods on the remaining datasets. On \cora and \cs datasets, we can consistently observe that \ours and \ourslight outperform all the graph augmentation baselines for \isolated and \cold nodes. Moreover, for \warm nodes, \ours can also perform on par or above baselines. On the \computers and \photos datasets, our methods generally achieve comparable or superior performance compared to the baselines, except in comparison to TuneUP. However, it is worth noting that both \ours and \ourslight exhibit more than 2$\times$ faster execution speed than TuneUP on these two datasets.

\subsection{Additional results under the inductive setting}
\label{sec:append_induc}

We further evaluate and present the effectiveness of our methods under the inductive setting on the remaining datasets in \cref{tab:product_appendix}. We can observe that both \ours and \ourslight consistently outperform GSage for \isolated, \cold, and \warm nodes. Compared to \ourslight, \ours is particularly beneficial for this inductive setting.
\begin{table}
\centering
\vspace{-0.2in}
\small
\caption{Performance in inductive settings (Remaining results of \cref{tab:product}). The best result is \textbf{bold}, and the runner-up is \underline{underlined}. Our methods consistently outperform GSage.}
\label{tab:product_appendix}
\scalebox{0.9}{
\begin{tabular}{l|c|ccc} 
\toprule
\multicolumn{2}{c|}{}&GSage &\ourslight &\ours \\\midrule
\multirow{4}{*}{\cora}&\isolated &\ms{43.64}{1.84} & \ms{\underline{45.31}}{0.83} & \ms{\textbf{46.06}}{0.66} \\
&\cold &\ms{60.06}{0.62} & \ms{\underline{60.46}}{0.91} & \ms{\textbf{61.94}}{2.22} \\
&\warm &\ms{60.59}{1.13} & \ms{\underline{60.95}}{1.40} & \ms{\textbf{62.53}}{1.23} \\
&\overall &\ms{57.23}{0.33} & \ms{\underline{57.65}}{0.82} & \ms{\textbf{59.24}}{1.02}\\\midrule
\multirow{4}{*}{\cs}&\isolated &\ms{74.34}{0.56} & \ms{\underline{75.42}}{0.36} & \ms{\textbf{77.80}}{0.68} \\
&\cold &\ms{75.75}{0.48} & \ms{\underline{77.02}}{0.65} & \ms{\textbf{81.33}}{0.60} \\
&\warm &\ms{82.55}{0.27} & \ms{\underline{83.52}}{0.67} & \ms{\textbf{83.55}}{0.50} \\
&\overall &\ms{81.00}{0.28} & \ms{\underline{82.01}}{0.59} & \ms{\textbf{82.70}}{0.52}\\\midrule
\multirow{4}{*}{\computers}&\isolated &\ms{66.81}{0.72} & \ms{\underline{67.03}}{0.51} & \ms{\textbf{69.82}}{0.63} \\
&\cold &\ms{64.17}{2.01} & \ms{\underline{65.10}}{1.76} & \ms{\textbf{66.36}}{0.69} \\
&\warm &\ms{68.76}{0.40} & \ms{\underline{68.78}}{0.39} & \ms{\textbf{70.49}}{0.41} \\
&\overall &\ms{68.54}{0.42} & \ms{\underline{68.59}}{0.39} & \ms{\textbf{70.40}}{0.42}\\\midrule
\multirow{4}{*}{\photos}&\isolated &\ms{68.29}{0.67} & \ms{\underline{69.60}}{0.75} & \ms{\textbf{70.46}}{0.53} \\
&\cold &\ms{63.02}{1.51} & \ms{\underline{64.25}}{1.31} & \ms{\textbf{68.49}}{2.39} \\
&\warm &\ms{70.17}{0.57} & \ms{\underline{71.05}}{0.70} & \ms{\textbf{71.61}}{0.81} \\
&\overall &\ms{69.92}{0.57} & \ms{\underline{70.84}}{0.63} & \ms{\textbf{71.47}}{0.77}\\
\bottomrule
\end{tabular}}
\vspace{-0.2in}
\end{table}

\begin{table}
\centering
\caption{Performance with different encoders (Remaining results of \cref{tab:ablation_encoder}), where the inner product is the decoder. The best result for each encoder is \textbf{bold}, and the runner-up is \underline{underlined}. Our methods consistently outperform the base models, particularly for \isolated and \cold nodes.}
\label{tab:ablation_encoder_appendix}
\scalebox{0.8}{
\begin{tabular}{l|c||ccc||ccc} 
\toprule
\multicolumn{2}{c||}{}&GAT &\ourslight &\ours &JKNet &\ourslight &\ours\\ \midrule
\multirow{4}{*}{\cora}&\isolated &\ms{25.61}{1.78} & \ms{\underline{30.73}}{2.54} & \ms{\textbf{36.83}}{1.76} & \ms{30.12}{1.02} & \ms{\underline{37.44}}{2.27} & \ms{\textbf{43.90}}{3.66} \\
&\cold &\ms{54.88}{0.84} & \ms{\underline{55.76}}{0.50} & \ms{\textbf{56.72}}{0.81} & \ms{59.56}{0.66} & \ms{\underline{61.93}}{1.64} & \ms{\textbf{62.89}}{1.43} \\
&\warm &\ms{\underline{55.31}}{1.14} & \ms{\textbf{55.36}}{1.28} & \ms{53.70}{1.26} & \ms{\underline{58.64}}{0.12} & \ms{\textbf{59.36}}{1.00} & \ms{57.67}{1.60} \\
&\overall &\ms{52.85}{0.91} & \ms{\textbf{53.58}}{0.80} & \ms{\underline{53.43}}{0.49} & \ms{56.74}{0.27} & \ms{\textbf{58.54}}{0.83} & \ms{\underline{58.40}}{1.33}\\\midrule
\multirow{4}{*}{\cs}&\isolated&\ms{33.74}{1.98} & \ms{\underline{34.77}}{0.90} & \ms{\textbf{41.76}}{2.99} & \ms{54.43}{1.77} & \ms{\underline{56.38}}{2.14} & \ms{\textbf{64.79}}{1.68} \\
&\cold &\ms{70.20}{0.47} & \ms{\underline{70.90}}{0.32} & \ms{\textbf{71.92}}{0.36} & \ms{73.97}{0.72} & \ms{\underline{76.64}}{0.38} & \ms{\textbf{77.77}}{0.43} \\
&\warm &\ms{\underline{78.39}}{0.28} & \ms{\textbf{78.67}}{0.33} & \ms{77.69}{0.89} & \ms{\underline{82.38}}{0.67} & \ms{\textbf{83.29}}{0.37} & \ms{79.20}{0.13} \\
&\overall &\ms{77.16}{0.24} & \ms{\textbf{77.49}}{0.30} & \ms{\underline{77.20}}{0.80} & \ms{\underline{81.35}}{0.62} & \ms{\textbf{82.41}}{0.32} & \ms{78.91}{0.13}\\\midrule
\multirow{4}{*}{\computers}&\isolated&\ms{12.04}{2.08} & \ms{\underline{16.84}}{2.34} & \ms{\textbf{17.17}}{2.22} & \ms{9.92}{3.07 } & \ms{\underline{23.81}}{2.02} & \ms{\textbf{25.50}}{1.32} \\
&\cold &\ms{53.60}{1.51} & \ms{\underline{53.62}}{1.00} & \ms{\textbf{53.65}}{2.35} & \ms{62.29}{1.08} & \ms{\underline{67.21}}{0.99} & \ms{\textbf{68.49}}{0.70} \\
&\warm &\ms{\textbf{60.19}}{1.19} & \ms{\underline{58.64}}{0.81} & \ms{58.55}{1.01} & \ms{69.96}{0.33} & \ms{\textbf{70.90}}{0.40} & \ms{\underline{70.66}}{0.25} \\
&\overall &\ms{\textbf{60.03}}{1.19} & \ms{58.50}{0.80} & \ms{\underline{58.77}}{1.93} & \ms{69.77}{0.32} & \ms{\textbf{70.78}}{0.40} & \ms{\underline{70.55}}{0.25}\\\midrule
\multirow{4}{*}{\photos}&\isolated&\ms{15.31}{3.46} & \ms{\underline{18.03}}{2.50} & \ms{\textbf{18.77}}{3.33} & \ms{12.77}{2.40} & \ms{\underline{19.44}}{1.31} & \ms{\textbf{20.56}}{1.61} \\
&\cold &\ms{43.11}{9.93} & \ms{\underline{43.40}}{9.61} & \ms{\textbf{44.21}}{9.25} & \ms{57.27}{2.06} & \ms{\underline{59.86}}{1.09} & \ms{\textbf{60.93}}{0.74} \\
&\warm &\ms{\underline{56.17}}{8.28} & \ms{\textbf{56.75}}{8.33} & \ms{56.10}{8.35} & \ms{68.35}{0.81} & \ms{\underline{69.56}}{0.69} & \ms{\textbf{69.60}}{0.50} \\
&\overall &\ms{55.91}{9.22} & \ms{\textbf{56.48}}{8.26} & \ms{\underline{55.93}}{8.28} & \ms{68.09}{0.82} & \ms{\underline{69.33}}{0.68} & \ms{\textbf{69.38}}{0.49}\\
\bottomrule
\end{tabular}}
\vspace{-0.2in}
\end{table}

\begin{table}
\centering
\vspace{0.15in}
\caption{Link prediction performance with MLP decoder (Remaining results of \cref{tab:ablation_decoder}), where GSage is the encoder. Our methods achieve better performance than the base model.}
\label{tab:ablation_decoder_appendix}
\scalebox{0.8}{
\begin{tabular}{l|c||ccc} 
\toprule
\multicolumn{2}{c||}{}&MLP-Dec. &\ourslight &\ours \\ \midrule
\multirow{4}{*}{\cora}&\isolated&\ms{16.83}{2.61} & \ms{\underline{37.32}}{3.87} & \ms{\textbf{38.41}}{1.22} \\
&\cold &\ms{58.83}{1.77} & \ms{\textbf{64.46}}{2.13} & \ms{\underline{64.02}}{1.02} \\
&\warm &\ms{\underline{58.84}}{0.86} & \ms{\textbf{61.57}}{0.98} & \ms{58.66}{0.61} \\
&\overall &\ms{55.57}{1.10} & \ms{\textbf{60.68}}{0.66} & \ms{\underline{58.93}}{0.25}\\\midrule
\multirow{4}{*}{\cs}&\isolated&\ms{5.60}{1.14 } & \ms{\underline{58.68}}{0.95} & \ms{\textbf{60.20}}{0.68} \\
&\cold &\ms{71.46}{1.08} & \ms{\underline{78.82}}{0.68} & \ms{\textbf{79.58}}{0.31} \\
&\warm &\ms{84.54}{0.32} & \ms{\textbf{85.88}}{0.22} & \ms{\underline{85.20}}{0.24} \\
&\overall &\ms{82.48}{0.32} & \ms{\textbf{84.96}}{0.25} & \ms{\underline{84.42}}{0.22}\\\midrule
\multirow{4}{*}{\computers}&\isolated&\ms{6.13}{3.63 } & \ms{\textbf{27.74}}{3.38} & \ms{\underline{26.70}}{3.98} \\
&\cold &\ms{62.56}{1.34} & \ms{\underline{62.60}}{3.38} & \ms{\textbf{63.35}}{3.64} \\
&\warm &\ms{69.72}{1.31} & \ms{\textbf{70.01}}{2.41} & \ms{\underline{68.43}}{2.50} \\
&\overall &\ms{69.53}{1.30} & \ms{\textbf{69.91}}{3.11} & \ms{\underline{68.30}}{2.51}\\\midrule
\multirow{4}{*}{\photos}&\isolated&\ms{6.34}{2.67}  & \ms{\underline{18.15}}{2.02} & \ms{\textbf{18.97}}{1.71} \\
&\cold &\ms{55.63}{6.21} & \ms{\textbf{56.13}}{6.36} & \ms{\underline{55.93}}{7.27} \\
&\warm &\ms{\underline{70.40}}{6.84} & \ms{\textbf{70.67}}{6.30} & \ms{69.97}{5.07} \\
&\overall &\ms{\underline{69.89}}{6.81} & \ms{\textbf{69.93}}{6.24} & \ms{69.69}{5.07}\\
\bottomrule
\end{tabular}}
\vspace{-0.2in}
\end{table}

\begin{table}[]
\centering
\small
\caption{Performance with GCN~\citep{kipf2016semi} and GT~\citep{dwivedi2020generalization} encoders, where the inner product is the decoder. The best result for each encoder is \textbf{bold}.}
\label{tab:ab_conv_encoder}
\scalebox{0.8}{
\begin{tabular}{ll|ccc|ccc}
\toprule
         &            & GCN                   & GCN+NodeDup(L)        & GCN+NodeDup           & GT           & GT+NodeDup(L)         & GT+NodeDup            \\ \midrule
\multirow{4}{*}{\cora}     & \isolated   & \ms{40.61}{3.52}          & \ms{42.93}{2.68}          & \textbf{\ms{46.71}{1.53}} & \ms{20.93}{2.46} & \textbf{\ms{38.82}{1.27}} & \ms{37.40}{1.53}          \\
         & Low-degree & \ms{63.86}{0.78}          & \textbf{\ms{64.63}{1.60}} & \ms{64.10}{1.37}          & \ms{58.59}{0.29} & \ms{61.16}{1.08}          & \textbf{\ms{61.39}{0.89}} \\
         & Warm       & \ms{60.59}{0.62}          & \textbf{\ms{61.31}{0.43}} & \ms{60.26}{0.70}          & \ms{58.14}{1.15} & \textbf{\ms{59.29}{0.84}} & \ms{59.07}{0.05}          \\
         & Overall    & \ms{60.16}{0.44}          & \textbf{\ms{61.02}{0.61}} & \ms{59.90}{0.89}          & \ms{55.40}{0.43} & \textbf{\ms{58.34}{0.19}} & \ms{58.18}{0.42}          \\ \midrule
\multirow{4}{*}{\citeseer} & Isolated   & \ms{45.56}{1.30}          & \ms{47.84}{0.94}          & \textbf{\ms{50.64}{1.10}} & \ms{36.84}{3.26} & \ms{51.46}{1.27}          & \textbf{\ms{52.34}{1.46}} \\
         & Low-degree & \ms{69.37}{0.36}          & \ms{70.15}{1.56}          & \textbf{\ms{71.13}{0.64}} & \ms{60.24}{1.18} & \ms{72.98}{1.54}          & \textbf{\ms{73.77}{1.03}} \\
         & Warm       & \textbf{\ms{74.68}{0.38}} & \ms{73.26}{0.97}          & \ms{72.93}{0.78}          & \ms{71.14}{1.47} & \ms{74.48}{1.08}          & \textbf{\ms{75.08}{0.63}} \\
         & Overall    & \ms{67.48}{0.42}          & \ms{67.47}{0.83}          & \textbf{\ms{67.67}{0.66}} & \ms{61.15}{1.57} & \ms{69.67}{1.10}          & \textbf{\ms{70.38}{0.86}} \\
 \bottomrule
\end{tabular}}
\end{table}

\subsection{Ablation study}
\label{append:ablation_study}
\subsubsection{Performance with various encoders and decoders}
For the ablation study, we further explored various encoders and decoders on the remaining datasets. The results are shown in \cref{tab:ablation_encoder_appendix} and \cref{tab:ablation_decoder_appendix}. From these two tables, we can observe that regardless of the encoders or decoders, both \ours and \ourslight consistently outperform the base model for \isolated and \cold nodes, which further demonstrates the effectiveness of our methods on cold nodes. Furthermore, \ourslight consistently achieves better performance compared to the base model for \warm nodes.  

Besides GSage, GAT and JKNet, we also conducted further experiments with convolutional-based GNNs, such as GCN~\citep{kipf2016semi} and GT(GraphTransformer)~\citep{dwivedi2020generalization}. The results are shown in~\cref{tab:ab_conv_encoder}. Our findings indicate that our methods can also improve performance when using GCN and GT as the encoder. However, since GCN uses the same matrix for both self-representations and neighbor representations, our methods only benefit from the supervision aspect. This leads to less pronounced performance improvements on cold nodes compared to using GT and GSage as the encoder. Specifically, NodeDup shows a 13.10\% improvement for GCN, 60.38\% for GT, and 29.79\% for GSage on isolated nodes. Moverover, NodeDup(L) on average improves GCN by 5.4\%, GT by 62.58\%, and GSage by 17.4\%. 

\begin{wraptable}{r}{0.5\linewidth}
\vspace{-0.3in}
\caption{Performance with SEAL~\citep{zhang2018link} on \cora and \citeseer datasets.}
\label{tab:seal}
\scalebox{0.75}{
\begin{tabular}{l|c|c|c}
\toprule
         &            & SEAL                  & SEAL + \ours         \\\midrule
\multirow{4}{*}{\cora}     & Isolated   & \ms{62.20}{1.06}          & \textbf{\ms{70.73}{0.61}} \\
         & Low-degree & \ms{66.80}{2.83}          & \textbf{\ms{67.70}{4.11}} \\
         & Warm       & \textbf{\ms{56.69}{2.36}} & \ms{54.87}{1.61}          \\
         & Overall    & \ms{60.60}{2.38}          & \textbf{\ms{60.89}{2.36}} \\\midrule
\multirow{4}{*}{\citeseer} & Isolated   & \ms{56.92}{5.53}          & \textbf{\ms{66.37}{1.01}} \\
         & Low-degree & \ms{64.13}{2.56}          & \textbf{\ms{65.54}{1.69}} \\
         & Warm       & \ms{58.81}{3.22}          & \textbf{\ms{60.73}{2.75}} \\
         & Overall    & \ms{60.18}{2.98}          & \textbf{\ms{63.35}{1.43}} \\
\bottomrule
\end{tabular}}
\vspace{-0.1in}
\end{wraptable}
\vspace{1in}
\subsubsection{Performance with SEAL~\citep{zhang2018link}}Considering our methods are flexible to integrate with GNN-based link prediction structures, we conduct the experiments on top of SEAL~\citep{zhang2018link} on the \cora and \citeseer datasets. The results are shown in~\cref{tab:seal}. We can observe that adding \ours on top of SEAL can consistently improve link prediction performance in the \isolated and \cold node settings on these two datasets. This further demonstrates the broad applicability of \ours in enhancing the performance of diverse GNN-based link prediction models.

\section{Implementation Details}
\label{sec:imple_detail}
In this section, we introduce the implementation details of our experiments. Our implementation can be found at \url{https://github.com/zhichunguo/NodeDup}.

\textbf{Parameter Settings.} We use 2-layer GNN architectures with 256 hidden dimensions for all GNNs and datasets. The dropout rate is set as 0.5. We report the results over 10 random seeds. Hyperparameters were tuned using an early stopping strategy based on performance on the validation set. We manually tune the learning rate for the final results. For the results with the inner product as the decoder, we tune the learning rate over range: $lr \in \{0.001, 0.0005, 0.0001, 0.00005 \}$. For the results with MLP as the decoder, we tune the learning rate over range: $lr \in \{0.01, 0.005, 0.001, 0.0005 \}$.

\textbf{Hardware and Software Configuration}
All methods were implemented in Python 3.10.9 with Pytorch 1.13.1 and PyTorch Geometric \citep{fey2019fast}. The experiments were all conducted on an NVIDIA P100 GPU with 16GB memory. 

\section{Limitations}
\label{sec:limitations}

In our work, \ours and \ourslight are specifically proposed for LP tasks. Although cold-start is a widespread issue in all graph learning tasks, our proposed methods might not be able to generalize to other tasks, such as node classification, due to their unique design. Furthermore, the two heterophily datasets we used for evaluation involve graphs where nodes with similar features are assigned different labels. Our methods may struggle on heterophilic graphs where connected nodes have dissimilar features, such as molecular networks, which are beyond the scope of this study.

\section{Ethics Statement}
\label{sec:ethics}

In this work, our simple but effective method enhances the link prediction performance on cold-start nodes, which mitigates the degree bias and advances the fairness of graph machine learning. It can be widely used and beneficial for various real-world applications, such as recommendation systems, social network analysis, and bioinformatics. We do not foresee any negative societal impact or ethical concerns posed by our method. Nonetheless, we note that both positive and negative societal impacts can be made by applications of graph machine learning techniques, which may benefit from the improvements induced by our work. Care must be taken, in general, to ensure positive societal and ethical consequences of machine learning.

\end{document}